%% file: main.tex
\documentclass[10pt,twocolumn,letterpaper]{article}

\usepackage[pagenumbers]{cvpr} 

\input{preamble}

\definecolor{cvprblue}{rgb}{0.21,0.49,0.74}
\usepackage[pagebackref,breaklinks,colorlinks,citecolor=cvprblue]{hyperref}
\usepackage{bbm}
\usepackage{multirow}
\usepackage{multicol}
\usepackage{amsmath,amsfonts}
\usepackage{graphicx}
\usepackage[utf8]{inputenc} 
\usepackage[T1]{fontenc}    

\usepackage{url}            
\usepackage{booktabs}       
\usepackage{amsfonts}       
\usepackage{nicefrac}       
\usepackage{microtype}      
\usepackage{xcolor}         
\newcommand{\B}{\boldsymbol}
\usepackage[noend]{algpseudocode}

\usepackage{algorithm}
\usepackage{array}
\renewcommand{\algorithmicrequire}{\textbf{Input:}}

\newcommand{\gray}[1]{\textcolor{gray}{{#1}}}
\usepackage{caption}
\def\paperID{10946} 
\def\confName{CVPR}
\def\confYear{2024}

\title{Black-box Adversarial Attacks Against Image Quality Assessment Models}

\author{Yu Ran\\
Guangzhou University\\
China\\
{\tt\small ranyu@e.gzhu.edu.cn}
\and
Ao-Xiang Zhang\\
Guangzhou University\\
China\\
{\tt\small zax@e.gzhu.edu.cn}
\and
MingJie Li\\
Guangzhou University\\
China\\
{\tt\small limingjie@gzhu.edu.cn}
\and
Weixuan Tang\\
Guangzhou University\\
China\\
{\tt\small tweix@gzhu.edu.cn}
\and
Yuan-Gen Wang\\
Guangzhou University\\
China\\
{\tt\small wangyg@gzhu.edu.cn}
}

\begin{document}
\maketitle
\input{sec/0_abstract}

\input{sec/1_intro}

\input{sec/2_related_work}
\input{sec/3_proposed_method}

\input{sec/4_experiment}

\input{sec/5_conclusion}
\input{sec/X_suppl}

{
    \small
    \bibliographystyle{ieeenat_fullname}
    \bibliography{main}
}


\end{document}

%% file: preamble.tex
\usepackage[dvipsnames]{xcolor}


%% file: sec/0_abstract.tex
\begin{abstract}
The goal of No-Reference Image Quality Assessment (NR-IQA) is to predict the perceptual quality of an image in line with its subjective evaluation. To put the NR-IQA models into practice, it is essential to study their potential loopholes for model refinement. 
This paper makes the first attempt to explore the black-box adversarial attacks on NR-IQA models. Specifically, we first formulate the attack problem as maximizing the deviation between the estimated quality scores of original and perturbed images, while restricting the perturbed image distortions for visual quality preservation. 
Under such formulation, we then design a Bi-directional loss function to mislead the estimated quality scores of adversarial examples towards an opposite direction with maximum deviation. 
On this basis, we finally develop an efficient and effective black-box attack method against NR-IQA models. Extensive experiments reveal that all the evaluated NR-IQA models are vulnerable to the proposed attack method. And the generated perturbations are not transferable, enabling them to serve the investigation of specialities of disparate IQA models.
\end{abstract}

%% file: sec/1_intro.tex
\section{Introduction}
\label{sec:intro}

\begin{figure}[t]
\renewcommand{\tabcolsep}{3pt}
\centering
\small
\begin{tabular}{cc}
\includegraphics[scale=0.145]{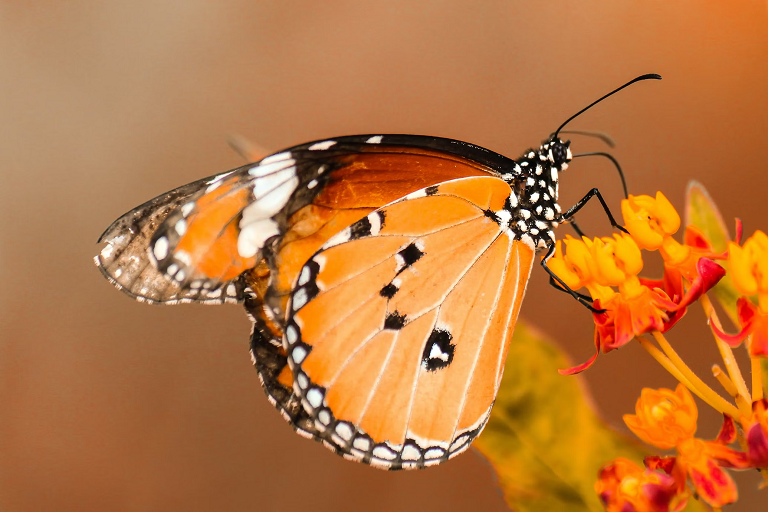}&
\includegraphics[scale=0.145]{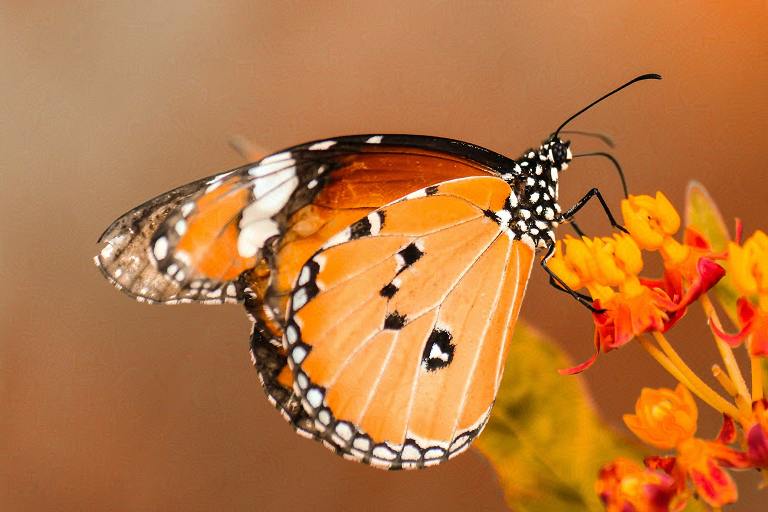}  \\
 (a) Original image  &  (b) Adversarial example \\
 Predicted quality score: 8.52 & Predicted quality score: 0.25 \\
\includegraphics[scale=0.145]{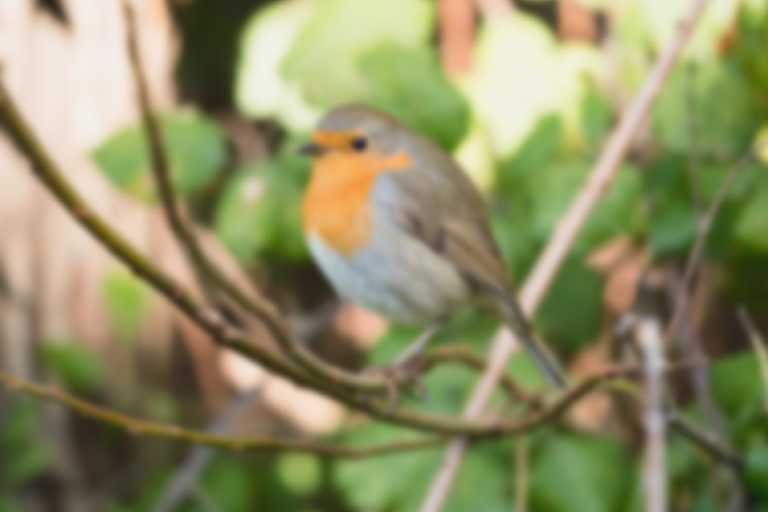} &
\includegraphics[scale=0.145]{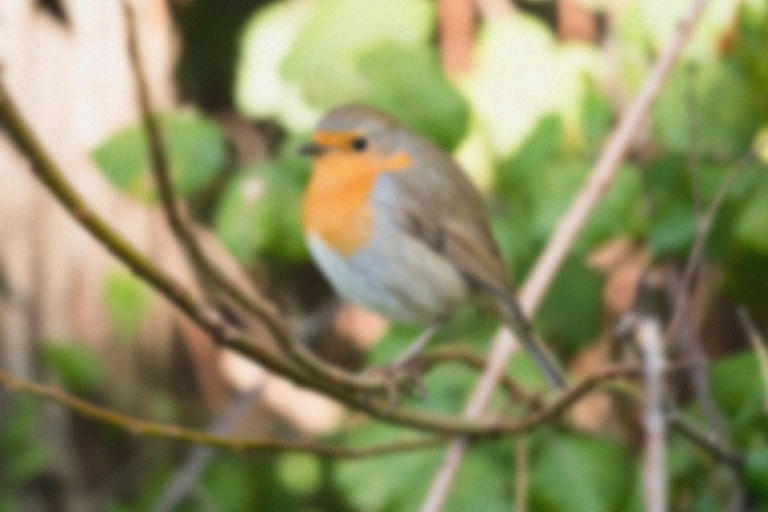}  \\
(c) Original image &  (d) Adversarial example \\ 
Predicted quality score: 3.44 & Predicted quality score: 9.72\\ 
\end{tabular}
\caption{\small An illustration for adversarial attacks against a DNN-based NR-IQA model. The quality of photos taken with a smartphone camera is becoming one of the key factors that users consider when purchasing a smartphone \cite{zhu2020multiple}. However, in real-world scenario, suppose that (a) and (c) are two photos taken with two different smartphone cameras, which will be sent to the third-party authoritative platform for quality assessment. (b) and (d) are their corresponding adversarial examples generated by attackers. If the original images are unfortunately replaced by adversarial examples. Then, we may see that the third-party platform, i.e., a DNNs-based NR-IQA model, would output very different quality scores for these two adversarial examples even their semantic information and visual qualities are preserved. These estimated quality results would significantly mislead the consumers' perception of the two smartphones.}
\label{fig: show}
\vspace{-2em}
\end{figure}

Image quality assessment (IQA) aims to quantify the perceptual quality of a digital image for providing users with good quality-of-experience. It has been an important research topic in image processing and understanding.
In the past, numerous IQA methods were proposed, which can mainly be classified into two categories, subjective and objective approaches. 
Subjective IQA \cite{ghadiyaram2015massive, hosu2020koniq} needs human evaluators to score a large number of images for estimation, which however is labor-intensive and time-consuming. 
In contrast, objective IQA \cite{ding2020image, golestaneh2022no} uses the subjective assessment data (e.g., the Mean Opinion Score (MOS)) for guidance and is able to automatically predict the perceptual qualities of images, attracting wide interest of researchers.
According to the availability of pristine reference images, objective IQA can roughly be categorized into full-reference IQA (FR-IQA) and no-reference IQA (NR-IQA). FR-IQA methods \cite{zhang2018unreasonable, ding2020image} rely on referenced undistorted images for quality prediction, which is impractical for scenarios where the pristine images are inaccessible. 
Therefore, substantial efforts are devoted to the NR-IQA paradigm \cite{gu2019no, ou2021novel, golestaneh2022no}, which plays a significant role in monitoring the image quality and optimizing the real-world image processing systems.

Due to the revolutionary successes of deep neural networks (DNNs) in various fields, DNNs-based NR-IQA approaches have been extensively studied and achieved state-of-the-art performance \cite{zhang2018blind, zhang2021uncertainty}. 
However, recent study indicates that DNNs-based models are vulnerable to adversarial examples \cite{zhang2022perceptual}. 
In other words, a well-trained DNNs-based NR-IQA model would incorrectly predict the quality score of an adversarial example that is generated by performing a small perturbation to the original image, i.e., adversarial attack (See Fig.~\ref{fig: show} for an illustration). 
Thus, to improve the robustness of NR-IQA models\footnote{For simplicity, we directly use NR-IQA to represent the DNNs-based NR-IQA in the following descriptions.}, it is indispensable to deeply understand their potential loopholes induced by adversarial perturbations.  
Nevertheless, it is highly nontrivial to craft visual semantics and quality-preserving attacks against NR-IQA models. The reason is that even perceptually invisible changes are highly likely to cause quality degradation. 
Recently, some studies \cite{zhang2022perceptual, shumitskaya2022universal} made efforts to conduct perceptual white-box attacks against NR-IQA models.
They showed that NR-IQA models are not perceptually robust to the white-box attacks. 
However, the crafted attacks under white-box setting are very impractical. This is because it is quite difficult for an attacker to completely have access to the victim IQA model information in real life. Zhang et al.~\cite{zhang2024vulnerabilities} investigated the vulnerability of video quality assessment (VQA) models by performing black-box adversarial attacks.     

In this paper, we make the first attempt to study the adversarial attacks against NR-IQA models in \textit{black-box} scenario, where the adversary is only allowed to query the victim models and receive the corresponding predictions. The main contributions of our work are \textit{fourfold}.
\begin{itemize}
\item We formulate the problem of black-box adversarial attacks on NR-IQA models with the goal of maximizing the deviation between the estimated qualities of original images and their adversarial examples. A distortion constraint is utilized to restrict the perturbation intensity for preserving the visual qualities of counterexamples. 

\item We design a Bi-directional loss function tailored for the proposed problem, which aims to directly mislead the estimated quality scores of adversarial examples towards the opposite direction of that of their original images. With this new loss function, we develop an efficient and effective black-box attack method against NR-IQA models. 



\item We conduct extensive experiments to evaluate four mainstream NR-IQA models under intra-model and inter-model black-box attacks on three benchmark IQA datasets. 
For better evaluation, we devise a new performance metric to measure the quality changes of adversarial examples.

\item Based on the experimental results, we achieve two significant observations: 1) The proposed attack method is capable of successfully fooling all four NR-IQA models. 2) The generated adversarial examples are not transferable, enabling them to serve the investigation of characteristics of disparate IQA models.

\end{itemize}

%% file: sec/2_related_work.tex
\section{Related Work}
In this section, we give out a detailed review for the related NR-IQA methods and the adversarial attacks. 

\subsection{NR-IQA Methods}
Early NR-IQA approaches focus on handling the specific distortions like JPEG compression and Gaussian blurring. Afterwards, general-purpose NR-IQA enjoys the popularity, which takes the overall perceptual quality of images into consideration. They can mainly be divided into natural scene statistics (NSS)-based and DNNs-based methods.  

\textbf{NSS-based IQA Methods.} NSS-based methods \cite{saad2012blind, mittal2012making, zhang2015feature, xu2016blind} rely on an assumption that the handcrafted features from natural images have some statistical regularities and the distorted images will break such regularities. Thus, this type of approach makes use of the handcrafted features with statistical regularities for visual quality prediction. 
A general NSS-based method is implemented by three steps: feature extraction, NSS modeling, and regression. Nevertheless, handcrafted feature-based methods still have limitations in modeling various real-life distortions. 

\textbf{DNNs-based IQA Methods.} Instead of using handcrafted features, DNNs-based methods \cite{kang2015simultaneous, bosse2017deep, bianco2018use, zhang2018blind, zhu2020metaiqa, su2020blindly, golestaneh2022no} directly use raw images as input and feed them into deep neural networks to automatically learn their perceptual representations for quality estimation. 
The work \cite{kim2018deep} utilizes DNNs to learn an objective error map, followed by a fine-tuning with handcrafted features to predict the subjective scores of images. 
The method \cite{zhang2018blind} considers the distortion level information and introduces a deep bilinear pooling network to enhance the prediction accuracy.  
Recently, Vision Transformer \cite{dosovitskiy2020image} is shown to be excellent at capturing the global feature dependencies for IQA tasks. 
The authors \cite{golestaneh2022no} employed Transformer to model the non-local dependency from multi-scale features extracted from DNNs, and then optimize the network by a loss combining the relative ranking and self-consistency information, to improve the robustness of NR-IQA.

\subsection{Adversarial Attacks} 
Adversarial attacks \cite{szegedy2013intriguing} were originally performed against DNNs-based classification models, with the goal of making the models misclassified.
The generated adversarial examples for attacking are semantically indistinguishable from their natural images to human eyes. Currently, there are two widely studied attack paradigms in literature: white-box and black-box attacks.

\textbf{White-box Attack.} 
This kind of attack needs to obtain all information about the victim model, and then uses gradient-based methods to produce adversarial examples. Representatives include the fast gradient sign method (FGSM)~\cite{goodfellow2014explaining}, iterative FGSM~\cite{kurakin2018adversarial}, projected gradient descent (PGD)~\cite{madry2017towards}. In addition, some useful strategies like Jacobian saliency map~\cite{papernot2016limitations} and Lagrangian relaxation~\cite{carlini2017towards} were introduced to enhance attack performance. 

\textbf{Black-box Attack.} The attack in this class differs from the white-box attack in that the adversary is only allowed to query the victim model and receive the associated predictions. 
Black-box attack is more practical in real-world applications, since it does not need internal information about the victim models. Thus, a large amount of methods were proposed in the past, which can mainly be divided into transfer-based~\cite{ demontis2019adversarial, qin2023training} and query-based \cite{andriushchenko2020square, ma2021simulating} attacks.
The former first trains a completely transparent substitute model locally and then crafts the adversarial example by attacking this
substitute model. Instead, the latter makes use of the
prediction information from victim model to construct the perturbations. 
However, black-box attack against IQA models still remains unsolved in literature. In this paper, we make the first attempt to explore this open issue. 



%% file: sec/3_proposed_method.tex
\section{Black-box Attacks on NR-IQA Models}
In this section, we first formulate black-box adversarial attack on NR-IQA model. Then, we introduce a new adversarial loss for perturbation optimization. Finally, we develop an efficient and effective black-box attack against NR-IQA models.

\subsection{Problem Formulation}
\label{sub:problem}

Let $\boldsymbol{x}\in[0,1]^d$ denote an original image, where its element scale varies from $0$ to $1$. $d = h \times w \times c$ is the image dimensionality, where $h$, $w$, and $c$ denote the height, width, and channel number of the image, respectively.
In IQA tasks, an image is usually associated with a Mean Opinion Score (MOS) that can be regarded as its true perceptual quality.
Let $y \in [\beta_{1}, \beta_{2}]$ denote the MOS value of an image.
Suppose that $\mathcal{F}$ denotes a well-trained victim NR-IQA model, and $\mathcal{F}(\B{x})\rightarrow \mathbb{R}$ represents the predicted quality score of the image $\B{x}$, which is mapped into the range from $\beta_{1}$ to $\beta_{2}$. $\mathcal{F}(\B{x})$ is usually close to the MOS of $\B{x}$.
A black-box adversarial attack on an NR-IQA model aims to modify a benign image $\B{x}$ to an adversarial example $\B{x}^\star$ under a distortion constraint, such that the deviation between $\mathcal{F}(\B{x})$ and $\mathcal{F}(\B{x}^\star)$ is maximized.
Formally, this problem is defined as:
\begin{equation}\label{eq:problem1}
\begin{aligned}
\B{x}^{\star} = \underset{\widetilde{\B{x}}}{\arg \max} \ \mathcal{D}(\mathcal{F}(\B{x}), \mathcal{F}(\widetilde{\B{x}})), \  s.t. \ \mathcal{T}(\B{x}, \widetilde{\B{x}}) < \rho,
\end{aligned}
\end{equation}
where $\widetilde{\B{x}}$ is the perturbed sample of $\B{x}$. 
$\mathcal{D}(\cdot,\cdot)$ denotes the deviation function between the estimated quality scores of $\B{x}$ and $\widetilde{\B{x}}$. 
In classification tasks, the cross-entropy function can be used to realize $\mathcal{D}$. However, for IQA problem, more suitable loss function needs to be designed.
$\mathcal{T}(\cdot,\cdot)$ represents the distortion metric between $\B{x}$ and $\widetilde{\B{x}}$, which can be evaluated by $\ell_{p}$-norm $\left \| \B{x} - \widetilde{\B{x}} \right \|_{p}$, or SSIM \cite{wang2004image}. Nevertheless, SSIM metric involves a gradient computation to restrict the image distortion, which is time-consuming. In this paper, we employ $\ell_\infty$-norm\footnote{$\ell_\infty$-norm is one of the widely studied $\ell_{p}$-norms for perceptual quality guarantee in adversarial learning.} to implement $\mathcal{T}$ due to its simplicity and effectiveness. 
To preserve the visual semantics and quality of an adversarial example, this distortion needs to be limited within a threshold $\rho$.

\subsection{Comparison to Attacks in Classification}
\label{sub:problem2}
Black-box attacks are widely investigated in classification task with the goal of misleading a well-trained classifier. The attacker aims to induce the victim model to produce an incorrect label for a perturbed image. However, the adversarial attacks on NR-IQA models greatly differ from that in classification by three points. 1) Classification tasks focus on predicting \textit{discrete} class labels for images correctly. In contrast, IQA problem is a regression task which produces \textit{continuous} quality scores (i.e., real numbers) for images with different qualities. 2) The attacks in classification should not change the visual semantics of original images. However, in IQA task, the attack needs to preserve both the visual semantics and qualities of images, which is much more challenging. 3) To perform a successful attack on IQA model, the adversary needs to mislead the victim model to generate a wrong output for an adversarial example while making the output have a large deviation with the correct prediction. This is much more complicated than the attacks in classification. Thus, an effective loss function as well as an efficient attack algorithm are the key to well address the proposed attack problem (i.e., Eq.~\ref{eq:problem1}), which will be introduced in detail below.    

\subsection{Bi-directional Adversarial Loss}
\label{sub: loss}

\begin{figure}[t]
\centering
\begin{subfigure}{0.48\linewidth}
\centering
\label{fig: white_MSE_loss}
\includegraphics[width=\linewidth]{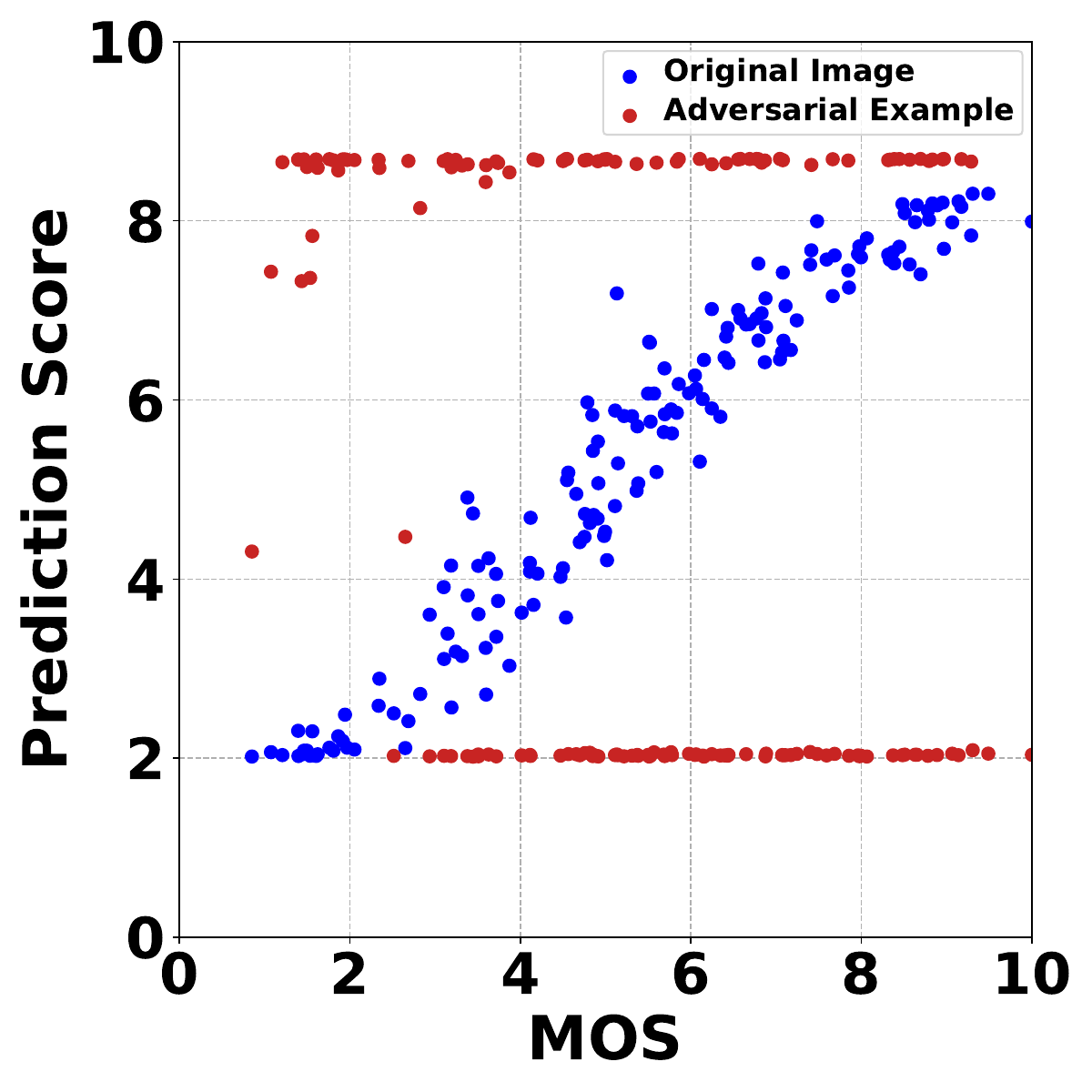}
\caption{MSE Loss}
\end{subfigure}
\begin{subfigure}{0.48\linewidth}
\centering
\label{fig: white_lccd_loss}
\includegraphics[width=\linewidth]{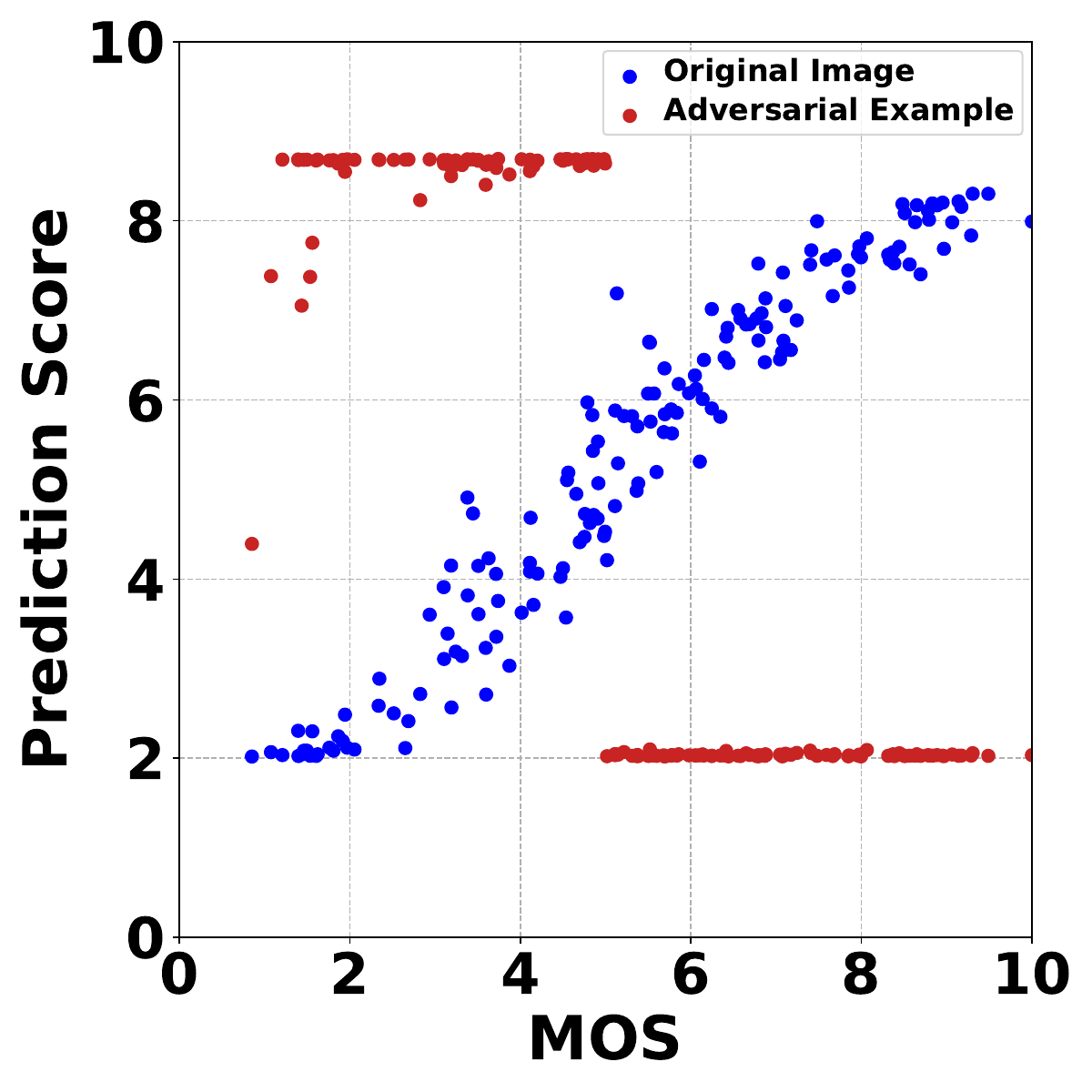}
\caption{Bi-directional Loss}
\end{subfigure}
\caption{\small The distributions of prediction scores using MSE loss and Bi-directional loss in white-box setting.} 
\label{fig: loss com}
\vspace{-0.2cm}
\end{figure}
\begin{figure}[tp]
\centering
\begin{subfigure}{0.48\linewidth}
\includegraphics[width=\linewidth]{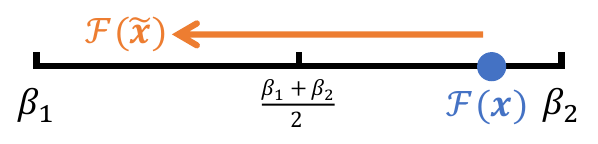}
\end{subfigure}
\begin{subfigure}{0.48\linewidth}
\includegraphics[width=\linewidth]{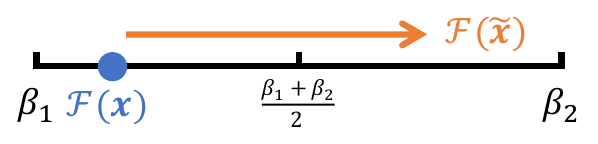}
\end{subfigure}
\caption{\small An illustration of the Bi-directional loss, where (a) is the case that $\mathcal{F}(\B{x})$ is greater than ($\beta_1+\beta_2)/2$, and (b) is the opposite case.}
\label{fig: Bi-di}
\vspace{-0.4cm}
\end{figure}

To design an effective loss function tailored for the above attack problem, we have the following considerations.
The frequently used margin-based \cite{carlini2017towards} and cross-entropy functions \cite{papernot2016limitations} in image classification task cannot directly be applied to IQA task due to the difference between their intrinsic problem attributes. Our goal is to maximize the deviation between the predicted qualities of an image $\B{x}$ and its counterexample $\widetilde{\B{x}}$. 
A feasible solution is using the mean square error (MSE) between $\mathcal{F}(\B{x})$ and $\mathcal{F}(\widetilde{\B{x}})$, which is a widely studied loss function in machine learning \cite{yang2022maniqa}. To investigate the effectiveness of MSE loss in IQA task, we conduct a white-box adversarial attack against UNIQUE \cite{zhang2021uncertainty} (a ``top-performing'' NR-IQA model) using the projected gradient descent (PGD) method~\cite{madry2017towards}.
The result is shown in Fig.~\ref{fig: loss com} (a).
We can see that MSE loss can only make the predicted quality scores of a portion of perturbed samples deviate from the MOSs of their associated original images, which is far from satisfactory.
The main reason is that maximizing MSE loss merely pushes the estimated scores of the perturbed images far away from that of their original images, which cannot control the directions of deviation, i.e., the predicted scores are probably smaller or larger than the MOS values. However, a successful attack is that if the MOS of an image falls into the large-value region (i.e., this is a high-quality image), then the predicted score of its adversarial example should be much smaller than the MOS value, and vice versa. This motivates us to develop a more effective loss function for the proposed problem.


In this paper, we design a Bi-directional loss function, which aims to mislead the predicted quality score of a perturbed image towards the opposite direction of that of its original image with maximal deviation. Specifically, our loss function with respect to an image $\B{x}$ and its perturbed version $\widetilde{\B{x}}$ is defined as:
\begin{equation}\label{eq: loss}
\mathcal{L}_{bd}(\mathcal{F}(\widetilde{\B{x}}), \mathcal{F}(\B{x})) = \left\{
\begin{aligned}
& \mathcal{F}(\widetilde{\B{x}}) ,\ \ \ \ \ \ \textrm{if} \ \ \mathcal{F}(\B{x}) > \frac{\beta_{1}+\beta_{2}}{2}  \\
& - \mathcal{F}(\widetilde{\B{x}}) ,\ \textrm{others}
\end{aligned}
\right.
\end{equation}
where $\beta_1$ and $\beta_2$ are the lower and upper bounds of the MOS values of all images within dataset, respectively.
$\mathcal{F}(\B{x})$ denotes the predicted quality score of image $\B{x}$, which is used to determine the optimization direction.
Note that $\widetilde{\B{x}}$ is usually initialized from $\B{x}$. Thus, the motivation of our Bi-directional loss is that through directly optimizing the deviation in the opposite direction, we are able to more easily maximize the difference between the estimated scores of an image and its adversarial example.
An illustration of our Bi-directional loss is given in Fig.~\ref{fig: Bi-di}.
Without loss of generality, we suppose that the victim NR-IQA model $\mathcal{F}$ performs well. 
If $\mathcal{F}(\B{x})$ is greater than $\frac{\beta_1+\beta_2}{2}$, then image $\B{x}$ is highly likely to be a high-quality image. In this case, it is desired to directly minimize $\mathcal{F}(\widetilde{\B{x}})$ for maximizing the deviation. 
For other case, it is desirable to maximize $\mathcal{F}(\widetilde{\B{x}})$, i.e., minimizing the negative $\mathcal{F}(\widetilde{\B{x}})$, for achieving the above purpose. In this manner, we are able to effectively realize the goal in our task. 
We perform an experiment to compare our loss function with the MSE loss function under the same configuration. The result is reported in Fig.~\ref{fig: loss com} (b). We can observe that our loss function is able to make the prediction scores of almost all perturbed samples deviate from that of their original images.

\begin{algorithm}[t]
\caption{Adversarial Example Generation Algorithm}
 \label{alg: alg2}
\begin{algorithmic}[1]
\renewcommand{\algorithmicrequire}{\textbf{Input:}}
\Require NR-IQA model $\mathcal{F}$, original image $\boldsymbol{x}$, its predicted score $\mathcal{F}(\boldsymbol{x})$, 
\renewcommand{\algorithmicrequire}{\textbf{Parameter:}}
\Ensure Adversarial example $\B{x^{\star}}$
\Procedure{Attack}{$\boldsymbol{x}, y, \mathcal{F}$}
\State Initialize a perturbed image $\widetilde{\B{x}}_{\rm opt} \gets \textsc{Init}(\B{x})$
\State Compute the loss ${J}_{\rm opt} \gets \mathcal{L}_{bd}(\mathcal{F}(\widetilde{\B{x}}_{\rm opt}), \mathcal{F}(\boldsymbol{x}))$
\For{each iteration $t=1, \ldots, T$}
	\State Generate a solution $\widetilde{\B{x}}_{t} \gets \textsc{Perturb} (\widetilde{\B{x}}_{\rm opt}, \B{x}, t)$
	\State 	Compute the new loss ${J}_{t} \gets \mathcal{L}_{bd}(\mathcal{F}(\widetilde{\B{x}}_{t}), \mathcal{F}(\boldsymbol{x}))$
	\If {$J_{t} < J_{\rm opt}$}
	\State Update the optimal loss $ J_{\rm opt} \gets J_{t}$
	\State Update the optimal solution $ \widetilde{\B{x}}_{\rm opt} \gets  \widetilde{\B{x}}_{t}$
	\EndIf
\EndFor
\State Achieve the adversarial example $\B{x^{\star}} \gets \widetilde{\B{x}}_{\rm opt}$
\State \Return $\B{x^{\star}}$
\EndProcedure
\end{algorithmic}
\end{algorithm}
\setlength{\textfloatsep}{0.6cm}
\begin{algorithm}[t]
\caption{Perturbation Algorithm}
 \label{alg: SG}
\begin{algorithmic}[1]
\renewcommand{\algorithmicrequire}{\textbf{Input:}}
\Require Original image $\boldsymbol{x}$, current optimal solution $\widetilde{\B{x}}_{\rm opt}$, current iteration $t$
\renewcommand{\algorithmicrequire}{\textbf{Parameter:}}
\Ensure A new perturbed image $\widetilde{\B{x}}_{t}$
\Procedure{Perturb}{$\widetilde{\B{x}}_{\rm opt}, \B{x}, t$}
\State Generate a modified sample $\B{x}_\Delta \gets \widetilde{\B{x}}_{\rm opt} - \B{x}$
\State Compute square size $s$ using Eq.~\ref{eq: s}
\State Produce $n$ square patches with size $s \times s \times c$
\For{each patch}
    \State $\B{p}_{1:s,1:s}$ $\gets$ sampling from $\{-3/255, 3/255\}$ 
    \State $\widetilde{\B{p}}$ $\gets$ repeat $\B{p}$ along the $c$ channels. 
    \State Randomly sample $l$ from $[0, h-s]$
    \State Randomly sample $r$ from $[0, w-s]$
    \State ${\B{x}_\Delta}_{\ l: l+s,r: r+s} \gets \widetilde{\B{p}}$ \gray{// replace for all channels}
\EndFor
\State  $\widetilde{\B{x}}_{t} \gets \B{x}_\Delta + \boldsymbol{x}$
\State Clip $\widetilde{\B{x}}_{t}$ into $[0, 1]$
\State \Return $\widetilde{\B{x}}_{t}$
\EndProcedure
\end{algorithmic}
\end{algorithm}
\par
\subsection{Adversarial Example Generation}
\label{sub: attack}
We design an efficient and effective random search-based black-box attack algorithm to generate the adversarial examples for Eq.~\ref{eq:problem1} using the above Bi-directional loss function. Its main idea is to generate a solution (i.e., a perturbed image) by performing a perturbation operation at each iteration, and accept this solution for the next round optimization if it improves the objective function. The technical details are summarized in Algorithm~\ref{alg: alg2}.

Specifically, our attack method mainly contains two processes: initialization and perturbation optimization. First, a perturbed image is initialized as $\widetilde{\B{x}}_{\rm opt}$, and the victim NR-IQA model $\mathcal{F}$ is queried with $\widetilde{\B{x}}_{\rm opt}$ to obtain the Bi-directional loss $J_{\rm opt}$. Since our objective function integrated with DNNs-based model is non-convex, a good initialization typically helps find better solutions, which will be introduced later.
Then, an iterative optimization process is carried out.
In each iteration, a new perturbed image $\widetilde{\B{x}}_t$ is generated using a perturbation algorithm. The new loss $J_t$ is accordingly achieved by querying $\mathcal{F}$ with $\widetilde{\B{x}}_t$. 
If $J_{t}$ is smaller than $J_{\rm opt}$, $\widetilde{\B{x}}_{t}$ is used to update the best solution found so far.
After $T$ times iterations, the best solution $\widetilde{\B{x}}_{\rm opt}$ is output as the final adversarial example ${\B{x}}^{\star}$. 

\textbf{Initialization.}
To enlarge the exploration at the beginning of optimization, we adopt a random strategy to initialize a perturbed image. Specifically, all the elements in original image $\B{x}$ (with size $h \times w \times c$) are added with values independently sampled from $\{-3/255, 3/255\}$ to generate $\widetilde{\B{x}}_{\rm opt}$. This strategy enables our method to achieve very good attack performance against all the victim models. We also empirically study the case without initialization, which can be seen in Subsection~\ref{subsec:ablation}.

\textbf{Perturbation Optimization.}
We develop a patch-based perturbation algorithm for optimization, which is presented in Algorithm~\ref{alg: SG}. Particularly, in the $t$-th iteration, a modified sample is first achieved by $\B{x}_\Delta = \widetilde{\B{x}}_{\rm opt} - \B{x}$. Then, $n$ square-shaped patches with size $s \times s \times c$ are generated. The square size $s$ is calculated by
\begin{equation}
\label{eq: s}
s = \sqrt{\gamma_{t} \times h \times w},
\end{equation}
where $\gamma_{0}$ is a hyper-parameter, and $\gamma_{t}$ is halved when $t \in \{10, 50, 200, 500, 1000, 2000, 4000, 6000, 8000\}$ if the maximal iteration $T$=$10000$, otherwise the decay positions for other $T$ are linearly rescaled by $\{10, 50, ..., 6000, 8000\}\times T / 10000$. The motivation of this decay scheme is that we first perform a broad search to explore the solution space and gradually exploit better solutions with local search. By this way, we can find better perturbations while significantly accelerating the search. $\gamma_{0}$ is used to control the initial square size.
The impact of $n$ and $\gamma_{0}$ will be investigated in Subsection~\ref{subsec:ablation}.
For each patch, the elements within the same channel are assigned with the same values independently sampled from $\{-3/255, 3/255\}$. 
Then, all square-shaped patches are used to replace the contents with the same size as the patches on $\B{x}_\Delta$ in randomly selected positions. 
Finally, a new perturbed image is yielded by $\widetilde{\B{x}}_{t} = \B{x}_\Delta + \boldsymbol{x}$, which is also clipped into $[0, 1]$.

\subsection{Difference with attacks on VQA Models}
\label{sub: VQA}
Some efforts~\cite{zhang2024vulnerabilities} have been made to conduct black-box attacks against video quality assessment (VQA) models. However, our method significantly differs from the attack approach in~\cite{zhang2024vulnerabilities} by two facts. (1) The loss function is designed for different purposes. In \cite{zhang2024vulnerabilities}, the loss function is devised to push the adversarial video's estimated quality score far away from its MOS value towards a specific boundary. It has two points that are different from ours. i) It relies on MOS to distinguish the qualities of original videos. But, we use the predicted scores of images to recognize the qualities, which is more practical. ii) It sets a specific boundary according to video quality for optimization, which may cause sub-optimal adversarial examples. In contrast, our \textit{Bi-directional} loss function aims to directly mislead the estimated quality scores of adversarial examples towards the \textit{opposite direction} of that of their original images, which is more consistent with the definition of adversarial attack. (2) The perturbation algorithms are different. Though both algorithms are based on a patch-based paradigm, the patch sampling in~\cite{zhang2024vulnerabilities} is based on a grid strategy without overlapping and the patch size is controlled by a hyper-parameter. Instead, our patch sampling is based on a random scheme, where the patch size is dynamically calculated using a decay mechanism, which is more efficient. Besides, the settings for patch values are also different. 

%% file: sec/4_experiment.tex
\section{Experiments}

In this section, we first set up the experiments, including the descriptions of IQA datasets and victim NR-IQA models, and the evaluation metrics for measuring the robustness of NR-IQA models and the performance of the proposed attack method. We then report the performance results for black-box attacks against the NR-IQA models accompanied by detailed analyses. 

\subsection{Experiment Setup}
\textbf{IQA Datasets.}
The experiments are conducted on three singly synthetically-distorted IQA datasets, including LIVE \cite{sheikh2006statistical}, CSIQ \cite{larson2010most}, and TID2013 \cite{ponomarenko2015image}.
We randomly sample $80\%$ images from each dataset to construct the training set and leave the remaining $20\%$ for testing.
For each attack, $50$ images are randomly sampled from testing set to evaluate its performance.
For a fair comparison, the MOS values from all datasets are uniformly rescaled into $[0, 10]$.


\begin{table*}[t]
\caption{\small The performance results of the four NR-IQA models under intra-model black-box attacks. The performance results of victim models before attacking are marked in gray.}
\vspace{-0.25cm}
\renewcommand\arraystretch{0.75}
\centering
\small
\begin{tabular}{c c ccc|ccc|ccc}
\toprule
 \multicolumn{1}{c}{\multirow{3}{*}{Model}} &  \multicolumn{1}{c}{\multirow{3}{*}{$T$}}  & \multicolumn{3}{c}{LIVE} & \multicolumn{3}{c}{CSIQ}& \multicolumn{3}{c}{TID2013}\\
\cmidrule(r){3-5}\cmidrule(r){6-8} \cmidrule(r){9-11}
&  & RGO & SRCC & PLCC &  RGO & SRCC & PLCC & RGO & SRCC & PLCC 
\\
\midrule
DBCNN & 10K 
& 0.30 & 0.32 \gray{(0.96)}& 0.36\gray{(0.97)} 
& 0.44 & -0.07 \gray{(0.89)} & -0.27 \gray{(0.92)} 
& 0.32 & 0.33 \gray{(0.86)} & 0.22 \gray{(0.92)}  \\ \cmidrule(r){1-11}
UNIQUE & 10K 
& 0.33 & 0.01 \gray{(0.97)} & -0.07 \gray{(0.97)}
& 0.45 & -0.31 \gray{(0.77)} & -0.37 \gray{(0.87)}
& 0.33 & -0.41 \gray{(0.80)} & -0.32 \gray{(0.81)}  \\ \cmidrule(r){1-11}
TReS &  10K 
& 0.34 & 0.14 \gray{(0.97)} & 0.20 \gray{(0.97)}
& 0.49 & -0.28 \gray{(0.91)} & -0.45 \gray{(0.91)} 
& 0.31 & -0.23 \gray{(0.93)} & -0.21 \gray{(0.93)} \\ \cmidrule(r){1-11}
LIQE & 10K 
& 0.32 & 0.39 \gray{(0.97)} & 0.32 \gray{(0.96)}
& 0.32 & 0.26 \gray{(0.87)} & 0.23 \gray{(0.87)} 
& 0.18 & 0.21 \gray{(0.80)} & 0.24 \gray{(0.80)} 
\\ 

\bottomrule
\end{tabular}	
\label{tab: black}
\vspace{-0.1cm}
\end{table*}

\begin{table*}[t]
\caption{\small The performance results of the NR-IQA models under inter-model black-box attacks. The performance results of victim models before attacking are marked in gray. D, U, T, and L denote the DBCNN, UNIQUE, TReS and LIQE models, respectively. $\rightarrow$ means that the adversarial examples generated by the left model are used as input for the right model.}
\vspace{-0.25cm}
\renewcommand\arraystretch{0.85}
\centering
\setlength\tabcolsep{6.8pt}
\begin{tabular}{ c ccc| ccc | ccc}
\toprule
 \multicolumn{1}{c}{\multirow{3}{*}{Transfer}} & \multicolumn{3}{c}{LIVE} & \multicolumn{3}{c}{CSIQ}& \multicolumn{3}{c}{TID2013}\\
\cmidrule(r){2-4}\cmidrule(r){5-7}\cmidrule(r){8-10}
 & RGO & SRCC & PLCC & RGO & SRCC & PLCC & RGO & SRCC & PLCC\\
\midrule
D $\rightarrow$ U 
& 0.04 & 0.96 \gray{(0.97)} & 0.96 \gray{(0.97)} 
& 0.04 & 0.77 \gray{(0.77)} & 0.83 \gray{(0.87)}  
& 0.04 & 0.80 \gray{(0.80)} & 0.81 \gray{(0.81)} \\
D $\rightarrow$ T 
& 0.02 & 0.97 \gray{(0.97)} & 0.97 \gray{(0.97)} 
& 0.07 & 0.85 \gray{(0.91)} & 0.85 \gray{(0.91)} 
& 0.02 & 0.94 \gray{(0.93)} & 0.93 \gray{(0.93)}  \\
D $\rightarrow$ L 
& 0.09 & 0.96 \gray{(0.97)} & 0.94 \gray{(0.96)} 
& 0.08 & 0.75 \gray{(0.87)} & 0.80 \gray{(0.87)} 
& 0.05 & 0.80 \gray{(0.80)} & 0.82 \gray{(0.80)} \\
U $\rightarrow$ D 
& 0.03 & 0.96 \gray{(0.96)} & 0.96 \gray{(0.97)}  
& 0.06 & 0.88 \gray{(0.89)} & 0.90 \gray{(0.92)} 
& 0.03 & 0.82 \gray{(0.86)} & 0.89 \gray{(0.92)} \\
U $\rightarrow$ T 
& 0.02 & 0.97 \gray{(0.97)} & 0.96 \gray{(0.97)}  
& 0.07 & 0.85 \gray{(0.91)} & 0.84 \gray{(0.91)} 
& 0.03 & 0.91 \gray{(0.93)} & 0.92 \gray{(0.93)}  \\
U $\rightarrow$ L 
& 0.11 & 0.95 \gray{(0.97)} &  0.93 \gray{(0.96)}  
& 0.08 & 0.76 \gray{(0.87)} & 0.79 \gray{(0.87)} 
& 0.06 & 0.81 \gray{(0.80)} & 0.83 \gray{(0.80)}  \\
T $\rightarrow$ D 
& 0.05 & 0.91 \gray{(0.93)} & 0.91 \gray{(0.93)} 
& 0.09 & 0.81 \gray{(0.88)} & 0.80 \gray{(0.91)} 
& 0.03 & 0.78 \gray{(0.85)} & 0.86 \gray{(0.90)}  \\ 
T $\rightarrow$ U 
& 0.06 & 0.91 \gray{(0.95)} & 0.91 \gray{(0.95)} 
& 0.06 & 0.53 \gray{(0.62)} & 0.58 \gray{(0.68)} 
& 0.05 & 0.74 \gray{(0.82)} & 0.74 \gray{(0.82)}  \\
T $\rightarrow$ L 
& 0.06 & 0.94 \gray{(0.96)} & 0.93 \gray{(0.94)} 
& 0.09 & 0.72 \gray{(0.86)} & 0.78 \gray{(0.87)}
& 0.05 & 0.69 \gray{(0.73)} & 0.76 \gray{(0.76)}  \\
L $\rightarrow$ D 
& 0.02 & 0.96 \gray{(0.96)} & 0.96 \gray{(0.97)} 
& 0.05 & 0.89 \gray{(0.89)} & 0.89 \gray{(0.92)} 
& 0.03 & 0.82 \gray{(0.86)} & 0.89 \gray{(0.92)}  \\
L $\rightarrow$ U
& 0.03 & 0.97 \gray{(0.97)} & 0.96 \gray{(0.97)} 
& 0.04 & 0.73 \gray{(0.77)} & 0.79 \gray{(0.87)} 
& 0.04 & 0.79 \gray{(0.80)} & 0.78 \gray{(0.81)}  \\
L $\rightarrow$ T 
& 0.02 & 0.97 \gray{(0.97)} & 0.97 \gray{(0.97)} 
& 0.06 & 0.88 \gray{(0.91)} & 0.89 \gray{(0.91)} 
& 0.02 & 0.92 \gray{(0.93)} & 0.92 \gray{(0.93)}  \\
\bottomrule
\end{tabular}	
\label{tab: transfer}
\vspace{-0.3cm}
\end{table*}

\textbf{Victim IQA Models.}
Four state-of-the-art DNNs-based NR-IQA models are selected as victim models to evaluate the black-box attack performance, including the CNN-based DBCNN~\cite{zhang2018blind}, Rank-based UNIQUE~\cite{zhang2021uncertainty}, Transformer-based TReS~\cite{golestaneh2022no} and LIQE~\cite{zhang2023blind}.
According to VQEG~\cite{video2000final}, the estimated quality scores output by all models are uniformly mapped into $[0, 10]$ using a monotonic logistic function~\cite{video2000final}.

\textbf{Evaluation Metrics.}
To conduct experiments for comparison, we design a new metric to evaluate the performance of black-box attacks against IQA models. This metric is defined as the average \underline{r}atio between the quality score changes of \underline{g}enerated adversarial examples and that of \underline{o}ptimal adversarial examples (RGO), compared to original images respectively. It is formally expressed as:
\begin{equation}\label{eq: ACR}
\text{RGO} = \frac{1}{k} \sum_{i = 1}^{k} \frac{\left |    \mathcal{F}(\B{x}^{\star}_{i}) - \mathcal{F}(\B{x}_{i})\right |}{\max \left \{ \beta_{2} - \mathcal{F}(\B{x}_{i}), \mathcal{F}(\B{x}_{i}) - \beta_{1} \right \} } ,
\end{equation}
where $k$ denotes the number of testing images, $\B{x}_{i}$ is the $i$-th original image, and $\B{x}^{\star}_{i}$ denotes its corresponding adversarial example. $\mathcal{F}(\cdot)$ stands for the quality score estimated by victim IQA model $\mathcal{F}$, and $\beta_1$ and $\beta_2$ represent the lower and upper bounds of MOS values, i.e., $1$ and $10$, respectively.
The denominator in Eq.~\ref{eq: ACR} actually denotes the maximum achievable change in quality score, which is regarded as the change caused by the optimal adversarial example.
Note that larger RGO means better attack performance on IQA model.
\par
In addition, we adopt two widely used metrics to measure the performance of the IQA models, the Spearman rank correlation coefficient (SRCC) and the Pearson linear correlation coefficient (PLCC). 
The former quantifies the monotonic relationship while the latter measures the linear correlation between the ground truth and the predicted quality scores, respectively.
\par
\textbf{Parameter Settings.}
The distortion threshold $\rho$ for an attack method depends on the resolutions of images and needs to satisfy the visual semantics and quality-preserving requirement as well. According to the empirical studies~\cite{chou1995perceptually, zhang2008just}, we set $\rho = 3/255$ for the adopted $\ell_\infty$-norm metric, which enables us to achieve the above purpose.
By default, the number of patches $n$ and the controlling factor $\gamma_{0}$ in our attack method are respectively set as $8$ and $0.09$ for LIQE model, and $2$ and $0.04$ for the other three models. Their impacts on the final attack performance are discussed later. 

\subsection{Performance Evaluation}
\label{sec: perform}
In this subsection, we report the performance evaluation results for the four NR-IQA models under two black-box attack modes. One is the intra-model attack, which uses the adversarial examples produced by the proposed attack on the corresponding IQA models.
The other is the inter-model attack that uses the adversarial examples originally generated by one IQA model to fool another IQA model. 
The results are presented in Tables~\ref{tab: black} and~\ref{tab: transfer}, based on which we have the following insightful observations. 
\begin{figure}[t]
\centering
\label{fig: subfig:a}
\begin{subfigure}{0.48\linewidth}
\centering
\includegraphics[width=\linewidth]{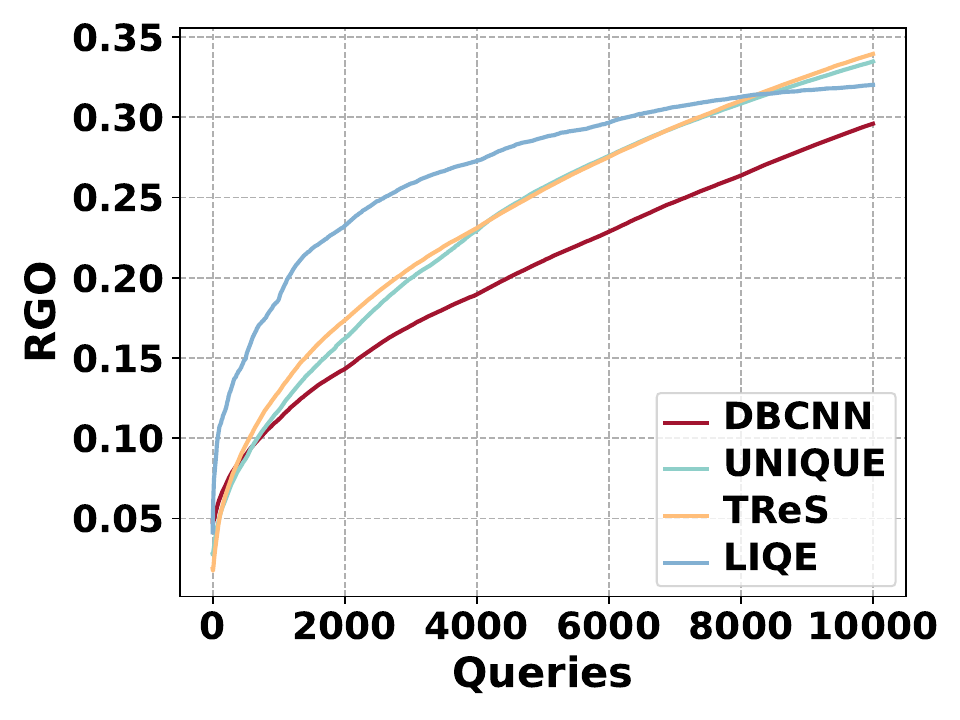}
\caption{LIVE}
\end{subfigure}
\begin{subfigure}{0.48\linewidth}
\centering
\includegraphics[width=\linewidth]{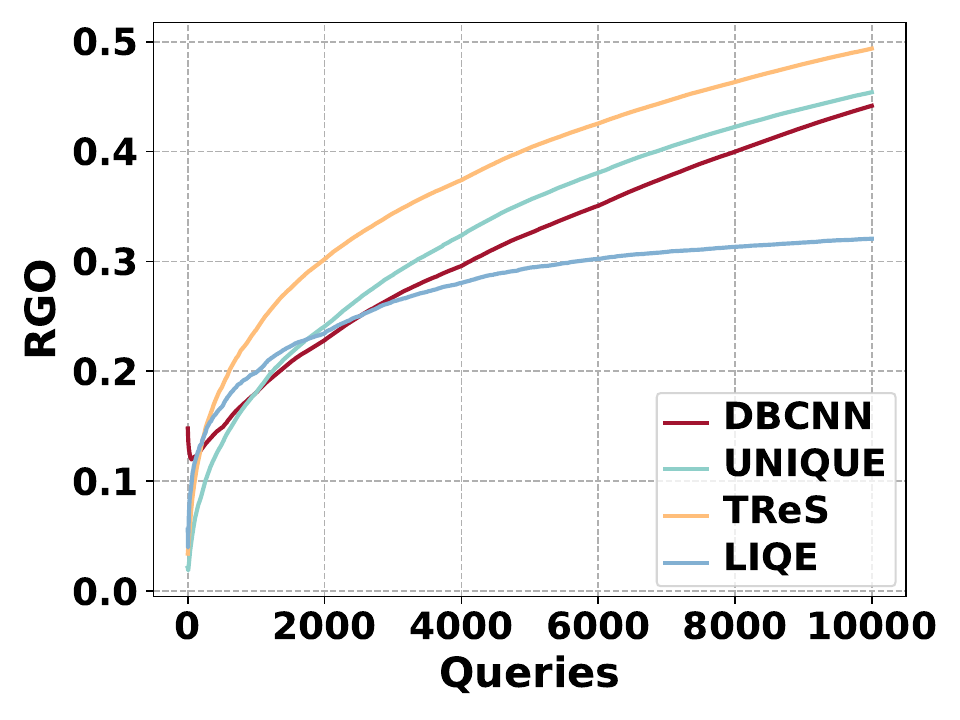}
\caption{CSIQ}
\end{subfigure}
\vspace{-0.2cm}
\caption{\small The curves of RGO w.r.t the number of queries by our attack method against UNIQUE model on LIVE and CSIQ datasets.}
\vspace{-0.4cm}
\label{fig: maindata}
\end{figure}
\begin{figure*}[t]
\begin{centering}   
\hspace{0.5cm}
\begin{subfigure}{0.31\linewidth}
\includegraphics[width=\linewidth]{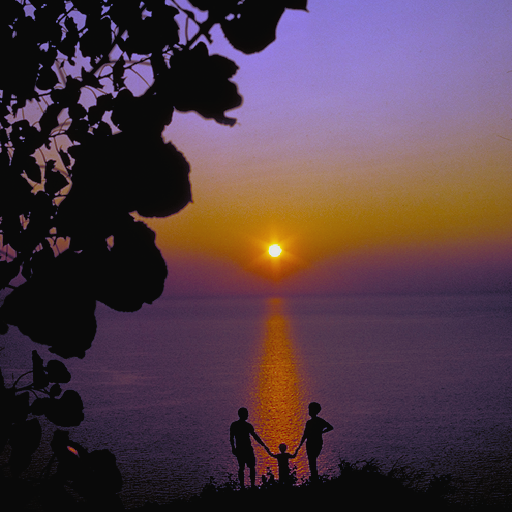}
\caption{Original Image \\ \ }
\end{subfigure}
\hspace{0.000mm}
\begin{subfigure}{0.15\linewidth}
\includegraphics[width=\linewidth]{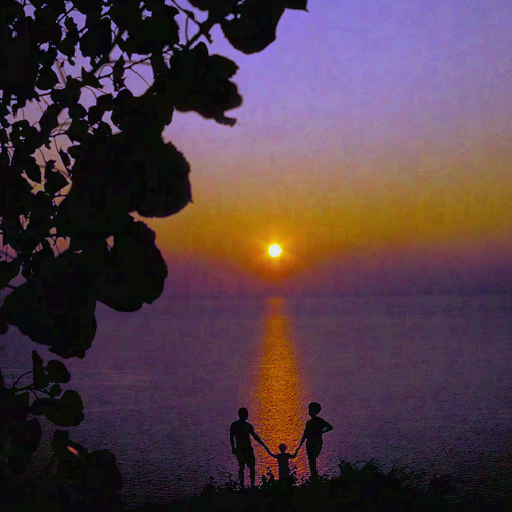} \\ \\ \vspace{-0.7cm}
\includegraphics[width=\linewidth]{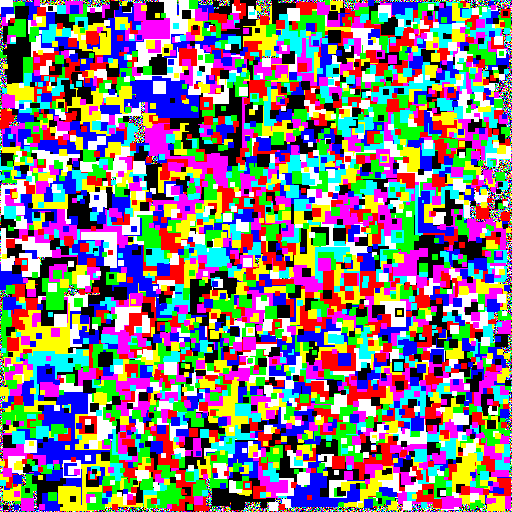}
\caption{\centering DBCNN \ \ \  \ \ \ \ \ \ \ \ \ \ \ \ \ \ \ 7.32 $\rightarrow$ 3.46}
\end{subfigure}
\hspace{0.000mm}
\begin{subfigure}{0.15\linewidth}
\includegraphics[width=\linewidth]{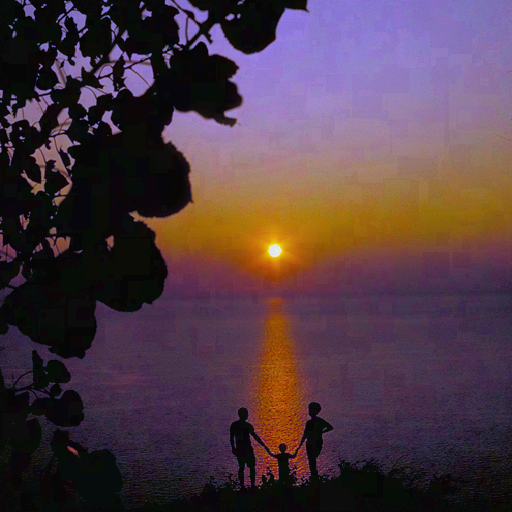} \\ \\ \vspace{-0.7cm}
\includegraphics[width=\linewidth]{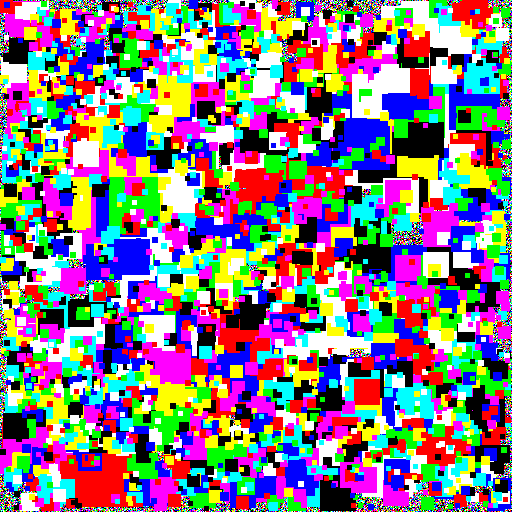}
\caption{\centering UNIQUE \ \ \ \ \ \ \ \ \ \ \ \ \ \ \ \ \ \ \ \ \ \ \ \  5.71 $\rightarrow$ 2.04}
\end{subfigure}
\hspace{0.000mm}
\begin{subfigure}{0.15\linewidth}
\includegraphics[width=\linewidth]{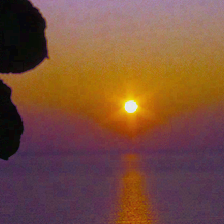} \\ \\ \vspace{-0.7cm}
\includegraphics[width=\linewidth]{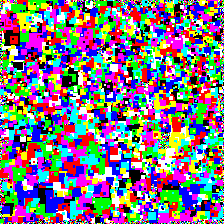}
\caption{TReS\\8.26 $\rightarrow$ 2.89}
\end{subfigure}
\hspace{0.000mm}
\begin{subfigure}{0.15\linewidth}
\includegraphics[width=\linewidth]{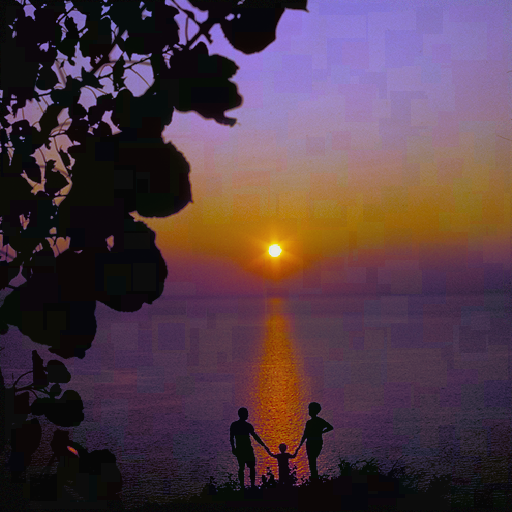} \\ \\ \vspace{-0.7cm}
\includegraphics[width=\linewidth]{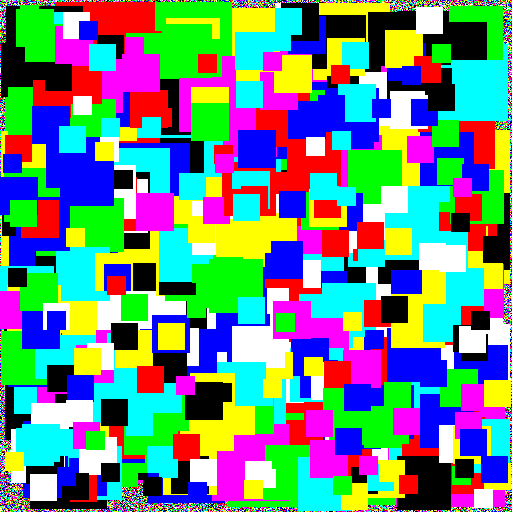}
\caption{LIQE\\6.52 $\rightarrow$ 2.66}
\end{subfigure}
\end{centering}

\vspace{-0.2cm}
\caption{\small The adversarial examples and their associated perturbations added to the original image (a) by attacking DBCNN (b), UNIQUE (c), TReS (d) and LIQE (e) models, respectively. $\rightarrow$ denotes the predicted quality score change between original image (a) and its adversarial example. Image (a) is from CSIQ dataset. Note that the image is center cropped to size $224\times224$ for TReS model due to its input constraint. Perturbations are scaled to $[0, 1]$ for visibility.}
\label{fig: vis}
\vspace{-0.2cm}
\end{figure*}
\par
\textbf{Intra-model Attack.}
The intra-model attack by our proposed attack method is able to successfully mislead all four NR-IQA models on all datasets.
Both SRCC and RLCC indicators of the victim models tremendously degrade compared with their original results. 
For example, both SRCC and PLCC values of TReS model before attacks are $0.91$ on CSIQ dataset. After intra-attacks, these two indicators degrade to $-0.28$ and $-0.45$, respectively. The average RGO that our attack method achieves on all the victim models on this dataset reaches $0.425$.
This suggests that our Bi-directional adversarial loss is able to maximize the difference between the predicted quality scores of an image and its adversarial example.
To provide a better understanding of the attack performance by our method, we plot in Fig.~\ref{fig: maindata} the RGO progression of adversarial examples w.r.t queries by the attacks against UNIQUE model on LIVE and CSIQ datasets. From the curves, we can see that the RGO by our method is steadily increasing even the number of queries reaches the maximum budget (i.e., $10000$) used in experiment. 
LIQE model shows relatively better robustness than the other models, which is mainly due to its multitask learning scheme for exploiting auxiliary knowledge from other tasks to serve the IQA task.  
Based on the above results, we conclude that DNNs-based NR-IQA models are not inherently perceptually robust to adversarial examples by black-box attacks. 
Our attack mechanism combined with the new loss function is quite suitable for the tasks of attacking NR-IQA models in black-box scenario.
To visualize our perceptual attack, we present in Fig.~\ref{fig: vis} the adversarial examples and their associated perturbations added to an original image from CSIQ dataset by attacking the four victim models, respectively.   
We can see that the proposed black-box attack is imperceptible. 

\begin{figure}[t]
\vspace{-0.2cm}
\begin{centering}      
\begin{subfigure}{0.48\linewidth}
\includegraphics[width=\linewidth]{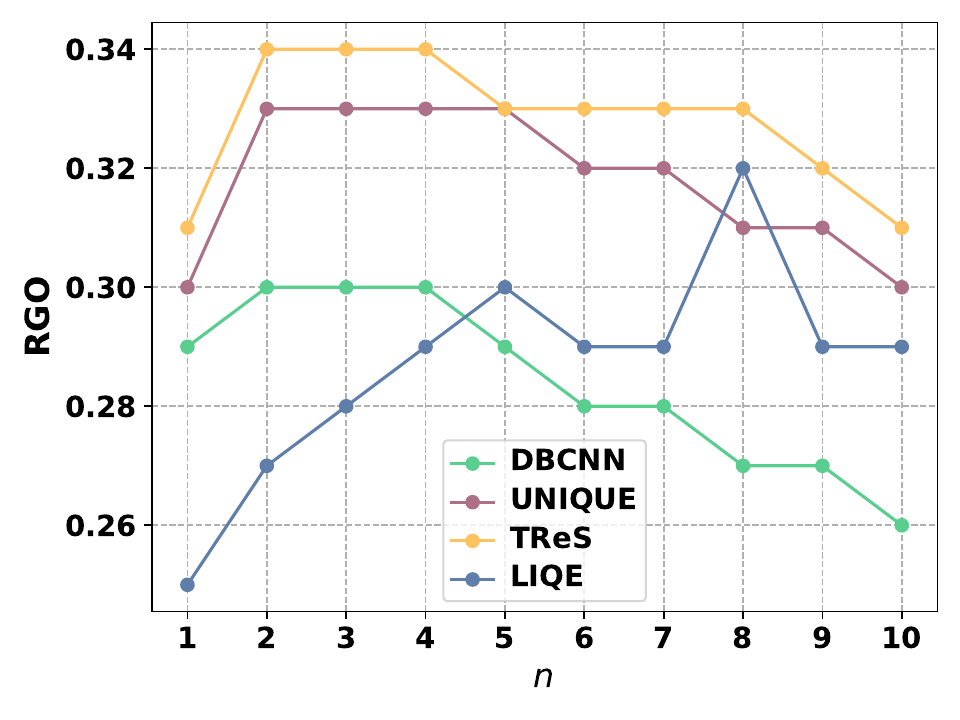}
\caption{LIVE}
\end{subfigure}
\begin{subfigure}{0.48\linewidth}
\includegraphics[width=\linewidth]{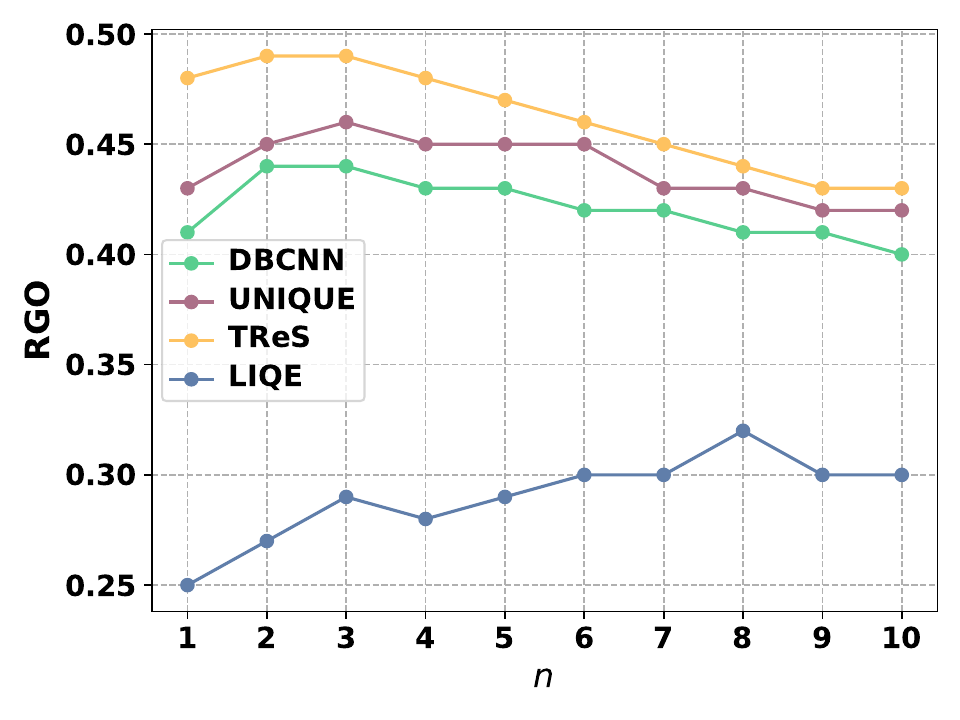}
\caption{CSIQ}
\end{subfigure}
\end{centering}
\vspace{-0.2cm}
\caption{\small The impact of the number of patches $n$ on attack performance for LIVE and CSIQ datasets.}
\label{fig: N}
\vspace{-0.3cm}
\end{figure}
\textbf{Inter-model Attack.} Compared with the former attack mode, we can see that all the evaluated NR-IQA models show to be very robust to the inter-model attacks.
For instance, using the perturbations by attacks on DBCNN model to fool the TReS model on LIVE dataset, both the SRCC and PLCC values remain the same as the original performances. The RGO performance of the attack is only $0.02$. The similar phenomenon also occurs in the attacks on other datasets. These results indicate the poor transferability
of perceptually imperceptible adversarial examples
against the IQA models in black-box setting. On one hand,
the adversarial examples crafted by the proposed attack method can naturally help investigate the different properties of different NR-IQA networks. On the other hand, this actually points out a new and interesting research direction for black-box attacks against NR-IQA models.

\begin{table*}[t]
\caption{\small The attack performances of the proposed method with and without initialization against the four victim models.}
\vspace{-0.2cm}
\small
\centering
\setlength\tabcolsep{3.1pt}
\renewcommand\arraystretch{0.75}
\begin{tabular}{ccc ccc|ccc|ccc}
\toprule
 \multicolumn{1}{c}{\multirow{3}{*}{Model}} & \multicolumn{1}{c}{\multirow{3}{*}{$T$}} & \multicolumn{1}{c}{\multirow{3}{*}{Initialization}}  & \multicolumn{3}{c}{LIVE} & \multicolumn{3}{c}{CSIQ}& \multicolumn{3}{c}{TID2013}\\
\cmidrule(r){4-6}\cmidrule(r){7-9}\cmidrule(r){10-12}
 & & & RGO & SRCC & PLCC & RGO & SRCC & PLCC & RGO & SRCC & PLCC \\
\midrule

\multirow{2}{*}{DBCNN} 
& \multirow{2}{*}{10K} & Without 
& 0.28 & 0.37 \gray{(0.96)} & 0.43 \gray{(0.97)} 
& 0.42 & 0.00 \gray{(0.89)} &- 0.15 \gray{(0.92)} 
& 0.31 & 0.35 \gray{(0.86)} & 0.21 \gray{(0.92)} \\
& & Uniform random 
& \bf0.30 & 0.32 \gray{(0.96)}& 0.36\gray{(0.97)} 
& \bf0.44 & -0.07 \gray{(0.89)} & -0.27 \gray{(0.92)} 
& \bf0.32 & 0.33 \gray{(0.86)} & 0.22 \gray{(0.92)} \\
\midrule
\multirow{2}{*}{UNIQUE} 
& \multirow{2}{*}{10K} & Without 
& 0.33 & 0.03 \gray{(0.97)} & -0.08 \gray{(0.97)} 
& 0.45 & -0.27 \gray{(0.77)} & -0.31 \gray{(0.87)} 
& 0.32 & -0.42 \gray{(0.80)} & -0.35 \gray{(0.81)} \\
& & Uniform random 
& \bf0.33 & 0.00 \gray{(0.97)} & -0.12 \gray{(0.97)}
& \bf0.45 & -0.32 \gray{(0.77)} & -0.38 \gray{(0.87)}
& \bf0.33 & -0.43 \gray{(0.80)} & -0.38 \gray{(0.81)} 
\\
\midrule
\multirow{2}{*}{TReS} 
& \multirow{2}{*}{10K} & Without 
& 0.34 & 0.14 \gray{(0.97)} & 0.19 \gray{(0.97)} 
& \bf0.50 & -0.29 \gray{(0.91)} & -0.46 \gray{(0.91)} 
& 0.31 & -0.22 \gray{(0.93)} & -0.15 \gray{(0.93)} 
\\
& & Uniform random 
& \bf0.34 & 0.14 \gray{(0.97)} & 0.20 \gray{(0.97)}
& 0.49 & -0.28 \gray{(0.91)} & -0.45 \gray{(0.91)} 
& \bf0.31 & -0.23 \gray{(0.93)} & -0.21 \gray{(0.93)} 
\\
\midrule
\multirow{2}{*}{LIQE} 
& \multirow{2}{*}{10K} & without 
& 0.25 & 0.61 \gray{(0.97)} & 0.67 \gray{(0.96)}
& 0.29 & 0.39 \gray{(0.87)} & 0.39 \gray{(0.87)} 
& 0.17 & 0.39 \gray{(0.80)} & 0.50 \gray{(0.80)} 
\\
& & Uniform random 
& \bf0.32 & 0.39 \gray{(0.97)} & 0.32 \gray{(0.96)}
& \bf0.32 & 0.26 \gray{(0.87)} & 0.23 \gray{(0.87)} 
& \bf0.18 & 0.21 \gray{(0.80)} & 0.24 \gray{(0.80)} 
\\ 
\bottomrule
\end{tabular}	
\label{tab: init}
\vspace{-0.1cm}
\end{table*}

\begin{table*}[t]
\caption{\small The attack performances of the proposed method using different losses.}
\vspace{-0.2cm}
\small
\centering
\setlength\tabcolsep{5pt}
\renewcommand\arraystretch{0.75}
\begin{tabular}{ccc ccc|ccc|ccc}
\toprule
 \multicolumn{1}{c}{\multirow{3}{*}{Model}} & \multicolumn{1}{c}{\multirow{3}{*}{$T$}} & \multicolumn{1}{c}{\multirow{3}{*}{Loss}}  & \multicolumn{3}{c}{LIVE} & \multicolumn{3}{c}{CSIQ}& \multicolumn{3}{c}{TID2013}\\
\cmidrule(r){4-6}\cmidrule(r){7-9}\cmidrule(r){10-12}
 & & & RGO & SRCC & PLCC & RGO & SRCC & PLCC & RGO & SRCC & PLCC \\
\midrule
\multirow{2}{*}{DBCNN} & \multirow{2}{*}{10K} & MSE 
& 0.29 & 0.49 \gray{(0.97)} & 0.47 \gray{(0.97)} 
& 0.40 & 0.07 \gray{(0.89)} & 0.05 \gray{(0.92)} 
& 0.31 & -0.24 \gray{(0.86)} & -0.37 \gray{(0.92)}\\
& & Bi-di 
& \bf0.30 & 0.32 \gray{(0.96)}& 0.36\gray{(0.97)} 
& \bf0.44 & -0.07 \gray{(0.89)} & -0.27 \gray{(0.92)} 
& \bf0.32 & 0.33 \gray{(0.86)} & 0.22 \gray{(0.92)} 
\\
\midrule
\multirow{2}{*}{UNIQUE} & \multirow{2}{*}{10K} &  MSE 
& 0.32 & 0.31 \gray{(0.97)} & 0.19 \gray{(0.96)} 
& 0.42 & -0.06 \gray{(0.77)} & -0.12 \gray{(0.87)} 
& 0.33 & 0.16 \gray{(0.80)} & -0.21 \gray{(0.81)}
\\
& & Bi-di 
& \bf0.33 & 0.01 \gray{(0.97)} & -0.07 \gray{(0.97)}
& \bf0.45 & -0.31 \gray{(0.77)} & -0.37 \gray{(0.87)}
& \bf0.33 & -0.41 \gray{(0.80)} & -0.32 \gray{(0.81)}   
\\
\midrule
\multirow{2}{*}{TReS} & \multirow{2}{*}{10K} &  MSE 
& 0.30 & 0.54 \gray{(0.97)} & 0.54 \gray{(0.97)}
& 0.44 & 0.39 \gray{(0.91)} & 0.20 \gray{(0.91)} 
& 0.27 & 0.35 \gray{(0.93)} & 0.23 \gray{(0.93)}
\\
& & Bi-di 
& \bf0.34 & 0.14 \gray{(0.97)} & 0.20 \gray{(0.97)}
& \bf0.49 & -0.28 \gray{(0.91)} & -0.45 \gray{(0.91)} 
& \bf0.31 & -0.23 \gray{(0.93)} & -0.21 \gray{(0.93)}
\\ 
\midrule
\multirow{2}{*}{LIQE} & \multirow{2}{*}{10K} &  MSE 
& 0.32 & 0.79 \gray{(0.97)} & 0.64 \gray{(0.96)}
& 0.29 & 0.40 \gray{(0.87)} & 0.40 \gray{(0.87)} 
& \bf0.19 & 0.69 \gray{(0.80)} & 0.63 \gray{(0.80)} 
\\
& & Bi-di
& \bf0.32 & 0.39 \gray{(0.97)} & 0.32 \gray{(0.96)}
& \bf0.32 & 0.26 \gray{(0.87)} & 0.23 \gray{(0.87)} 
& 0.18 & 0.21 \gray{(0.80)} & 0.24 \gray{(0.80)} 
\\
\bottomrule
\end{tabular}
\label{tab: loss}
\vspace{-0.3cm}
\end{table*}

\begin{table*}[h]
\caption{\small Performance comparison between Zha2023 \cite{zhang2024vulnerabilities} and Ours on different datasets.}
\vspace{-0.3cm}
\footnotesize
\centering
\setlength\tabcolsep{5.5pt}
\begin{tabular}{ccc ccc|ccc|ccc}
\toprule
\multicolumn{1}{c}{\multirow{3}{*}{Model}} & \multicolumn{1}{c}{\multirow{3}{*}{$T$}} & \multicolumn{1}{c}{\multirow{3}{*}{Loss}}  & \multicolumn{3}{c}{LIVE} & \multicolumn{3}{c}{CSIQ}& \multicolumn{3}{c}{TID2013}\\
\cmidrule(r){4-6}\cmidrule(r){7-9}\cmidrule(r){10-12}
 & & & RGO & SRCC & PLCC & RGO & SRCC & PLCC & RGO & SRCC & PLCC \\
\midrule
\multirow{2}{*}{DBCNN} & \multirow{2}{*}{10K} & Zha2023 
& 0.21 & 0.52 \gray{(0.97)} & 0.58 \gray{(0.97)} 
& 0.36 & -0.37 \gray{(0.89)} & -0.47 \gray{(0.92)} 
& 0.22 & -0.49 \gray{(0.86)} & -0.27 \gray{(0.92)}\\
& & Ours 
& \bf 0.30 & 0.29 \gray{(0.96)}& 0.32\gray{(0.97)} 
& \bf0.43 & -0.03 \gray{(0.89)} & -0.17 \gray{(0.92)} 
& \bf0.31 & 0.35 \gray{(0.86)} & 0.25 \gray{(0.92)} \\
\midrule
\multirow{2}{*}{UNIQUE} & \multirow{2}{*}{10K} &  Zha2023 
& 0.15 & 0.72 \gray{(0.97)} & 0.77 \gray{(0.96)} 
& 0.24 & 0.10 \gray{(0.77)} & 0.21 \gray{(0.87)} 
& 0.17 & 0.03 \gray{(0.80)} & 0.06 \gray{(0.81)}\\
& & Ours 
& \bf0.33 & 0.01 \gray{(0.97)} & -0.07 \gray{(0.97)}
& \bf0.45 & -0.31 \gray{(0.77)} & -0.37 \gray{(0.87)}
& \bf0.33 & -0.41 \gray{(0.80)} & -0.32 \gray{(0.81)}  \\
\midrule
\multirow{2}{*}{TReS} & \multirow{2}{*}{10K} &  Zha2023 
& 0.17 & 0.73 \gray{(0.97)} & 0.79 \gray{(0.97)}
& 0.24 & 0.21 \gray{(0.91)} & 0.23 \gray{(0.91)} 
& 0.14 & 0.46 \gray{(0.93)} & 0.59 \gray{(0.93)}
\\
& & Ours 
& \bf0.34 & 0.16 \gray{(0.97)} & 0.21 \gray{(0.97)}
& \bf0.50 & -0.30 \gray{(0.91)} & -0.46 \gray{(0.91)} 
& \bf0.31 & -0.23 \gray{(0.93)} & -0.23 \gray{(0.93)}
\\ 
\midrule
\multirow{2}{*}{LIQE} & \multirow{2}{*}{10K} &  Zha2023 
& 0.01 & 0.96 \gray{(0.97)} & 0.95 \gray{(0.96)}
& 0.01 & 0.85 \gray{(0.87)} & 0.87 \gray{(0.87)} 
& 0.01 & 0.80 \gray{(0.80)} & 0.79 \gray{(0.80)} 
\\
& & Ours 
& \bf0.30 & 0.50 \gray{(0.97)} & 0.44 \gray{(0.96)}
& \bf0.30 & 0.37 \gray{(0.87)} & 0.37 \gray{(0.87)} 
& \bf0.17 & 0.37 \gray{(0.80)} & 0.48 \gray{(0.80)} \\ 
\bottomrule
\end{tabular}	
\label{tab: vqa}
\vspace{-0.3cm}
\end{table*}

\begin{figure}[t]
\vspace{-0.2cm}
\begin{centering}      
\begin{subfigure}{0.48\linewidth}
\includegraphics[width=\linewidth]{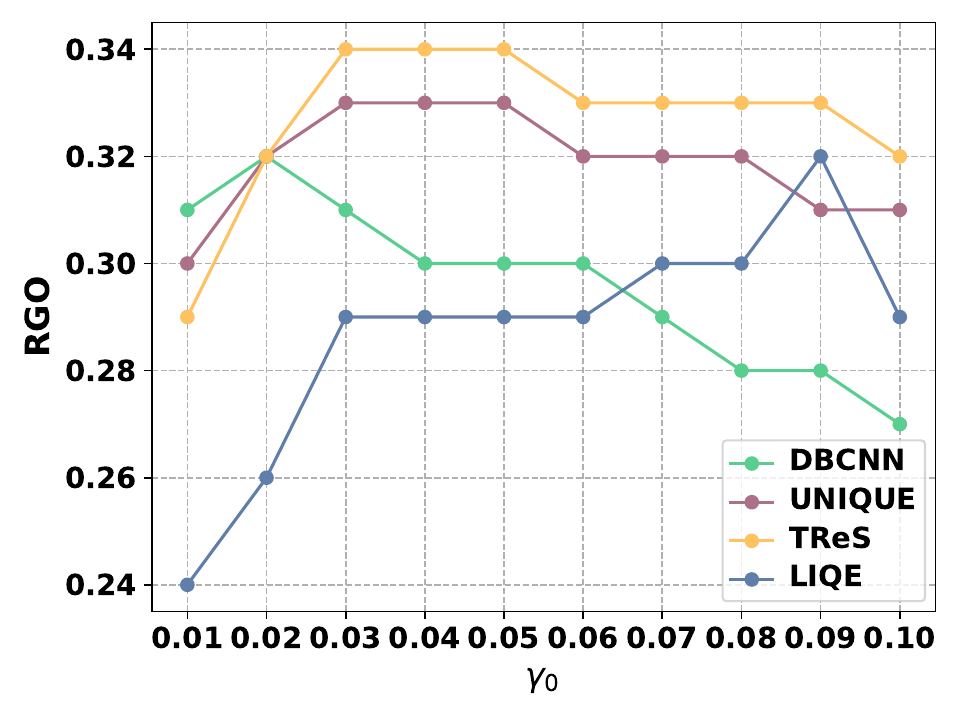}
\caption{LIVE}
\end{subfigure}
\begin{subfigure}{0.48\linewidth}
\includegraphics[width=\linewidth]{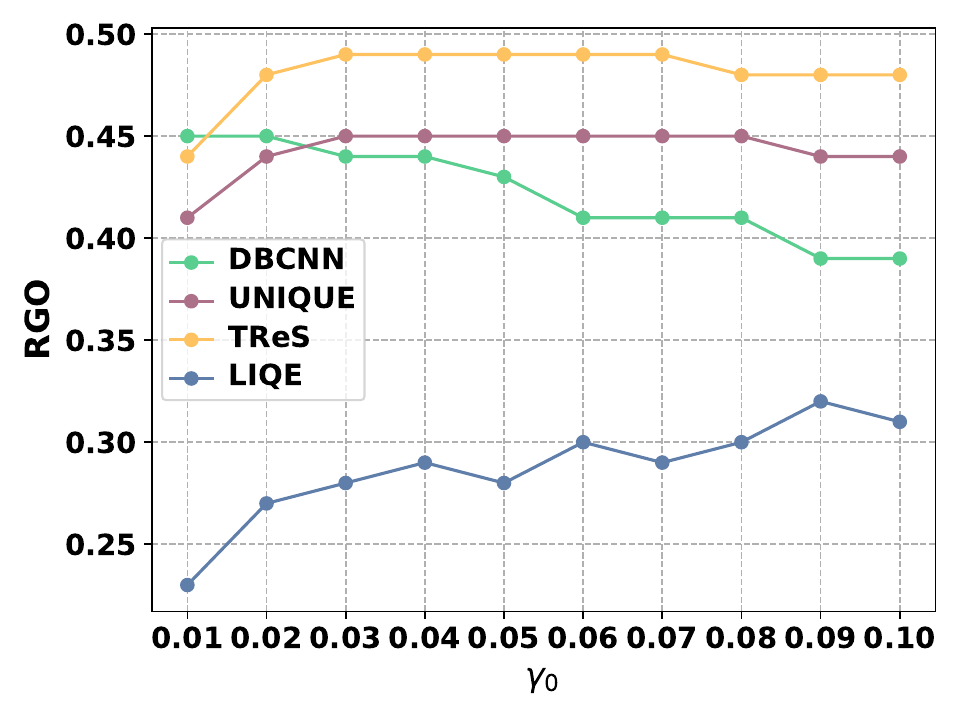}
\caption{CSIQ}
\end{subfigure}
\end{centering}
\vspace{-0.2cm}
\caption{\small The impact of controlling factor $\gamma_{0}$ on the attack performance for LIVE and CSIQ datasets.}
\label{fig: gamma0}
\vspace{-0.3cm}
\end{figure}

\textbf{Comparison to attacks tailored for VQA models.} We appropriately modify the black-box attack method in \cite{zhang2024vulnerabilities} and apply it into attacking NR-IQA models on different datasets, denoted as Zha2023. The experimental results are reported in Table~\ref{tab: vqa}, showing that our attack outperforms \cite{zhang2024vulnerabilities} on all victim models. We observe that \cite{zhang2024vulnerabilities} can hardly succeed in attacking LIQE model which adopts a multitask learning scheme. 
\subsection{Hyper-parameter and Ablation Study}
\label{subsec:ablation}
In this subsection, we conduct experiments to study the impact of hyper-parameters, as well as the effectiveness of the initialization strategy and the proposed Bi-directional loss function in our attack method.  

\textbf{Hyper-parameter.} There are two hyper-parameters need to be studied in our method, i.e., the number of patches $n$ and the controlling factor $\gamma_{0}$. Note that $\gamma_{0}$ is used to control the initial square size of patches. 
We plot the curves of RGO w.r.t different $n$ and $\gamma_{0}$ for the four NR-IQA models on LIVE and CSIQ datasets in Figs.~\ref{fig: N} and ~\ref{fig: gamma0}, respectively, where $n$ changes from $1$ to $10$ and $\gamma_{0}$ varies from $0.01$ to $0.10$.
From the Fig.~\ref{fig: N}, we observe that our method achieves the best performance when $n$ equals $8$ for LIQE model and $2$ for the other three models.
Based on Fig.~\ref{fig: gamma0}, our method obtains the best RGO for $\gamma_{0}=0.09$ on LIQE and $0.04$ on the other models. 
Therefore, we choose this parameter setting for all the experiments. 


\textbf{Initialization.}
We investigate how the random initialization influences the attack performance. The attack performances of our method with and without initialization against the four victim models on all datasets are reported in Table~\ref{tab: init}. We can observe that the performance difference between these two cases is not very prominent. However, the variant with initialization has a slightly higher RGO performance on most models, especially on LIQE. 

\begin{figure}[t]
\centering
    \begin{subfigure}{0.48\linewidth}
    \centering
    \includegraphics[width=\linewidth]{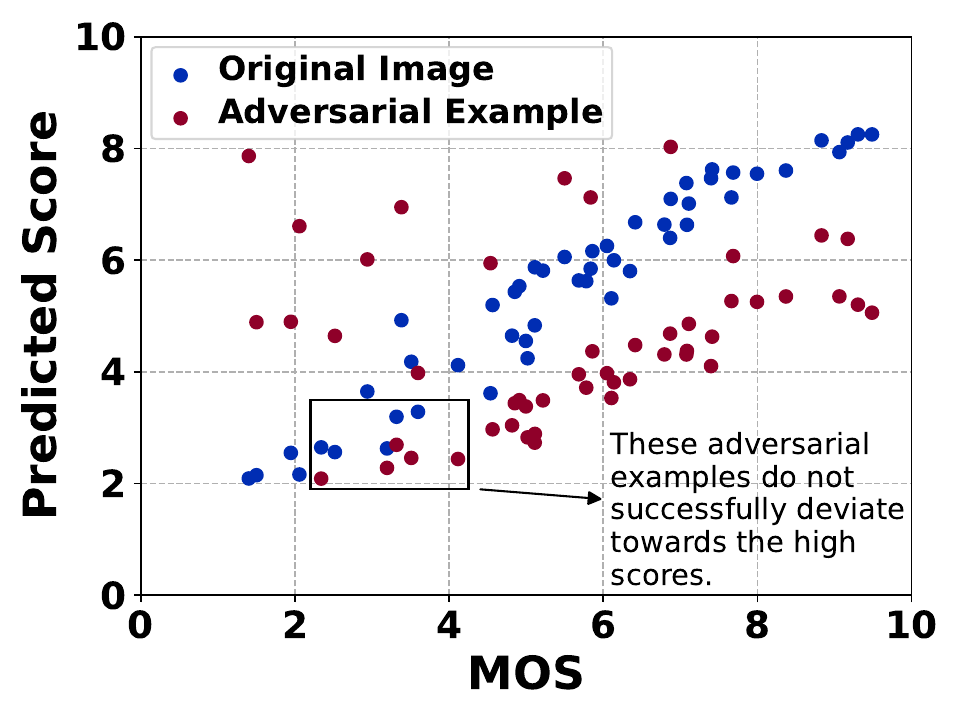}
    \caption{MSE loss}
    \label{fig: bl_MSE_loss}
\end{subfigure}
\begin{subfigure}{0.48\linewidth}
    \centering
    \includegraphics[width=\linewidth]{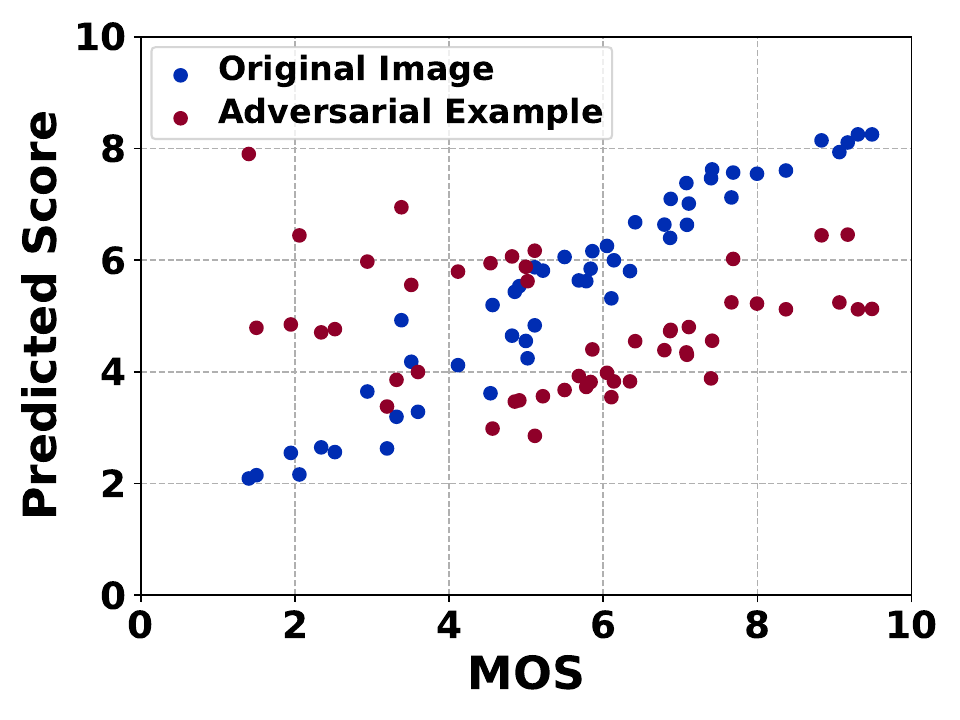}
    \caption{Bi-directional loss}
    \label{fig: bl_lccd_loss}
\end{subfigure}
\vspace{-0.2cm}
\caption{\small The distributions of prediction scores by our method using MSE and Bi-directional losses against UNIQUE model on LIVE dataset.} 
\label{fig: loss}
\vspace{-0.1cm}
\end{figure}
\textbf{Adversarial Loss.}
We study the effectiveness of the proposed Bi-directional loss function by comparing it with the Mean Square Error (MSE) loss in black-box setting on all datasets. 
We report the performances of these two loss functions using the proposed attack method in Table~\ref{tab: loss}. 
From the results, we can see that the proposed loss function outperforms the MSE loss on all victim models.
For example, the original SRCC and PLCC of TReS model on CSIQ dataset are both $0.91$. After perturbation, the proposed loss function is capable of making these two indicators degrade to -$0.28$ and -$0.45$, respectively, achieving a high attack performance of $0.49$ RGO. However, the MSE loss only results in a small degradation with $0.39$ SRCC and $0.20$ PLCC, as well as a worse performance with $0.44$ RGO.

In Fig.~\ref{fig: loss}, we visualize the distributions of prediction scores using MSE loss and Bi-directional loss, respectively, where our attack method is employed to attack UNIQUE model on LIVE dataset. 
It shows that the proposed loss has the ability to enable the prediction scores of most adversarial examples to deviate from that of their original images, while the MSE loss can hardly achieve this aim. As illustrated in Fig.~\ref{fig: loss} (a), the quality scores of adversarial examples in the rectangle box do not successfully deviate towards the high scores.

%% file: sec/5_conclusion.tex
\section{Conclusion}
In this paper, we take the first step towards investigating the black-box adversarial attacks against the NR-IQA models. We first formulate this new problem by borrowing the methodologies from both adversarial attacks and IQA models. 
To address this problem, we then design an effective black-box attack method using a new Bi-directional loss function for optimization. We also devise a new attack performance metric for evaluation.
Extensive experiments demonstrate that the DNNs-based NR-IQA models are not perceptually robust to the adversarial examples crafted by the attacks in black-box scenario. Furthermore, our study shows that the generated perturbations by one NR-IQA model are not able to successfully mislead other models, which contain useful information to investigate the specialities of different models. We believe that this exploration of black-box attacks on NR-IQA models can naturally help put the NR-IQA methods into practice.     

%% file: sec/X_suppl.tex
\newpage
\setcounter{section}{0}
\maketitlesupplementary
\renewcommand\thesection{\Alph{section}}

\label{sec: stability}
\begin{table}[t]
\renewcommand{\thetable}{S1}
\begin{minipage}[c]{1\textwidth}
\caption{\small The performance results of our attack method using different distortion threshold $\rho$. The performance results of victim models before attacking are marked in gray.}
\vspace{-0.25cm}
\centering
\setlength\tabcolsep{8pt}
\scriptsize  
\begin{tabular}{c c c ccc|ccc|ccc}
\toprule
\multicolumn{1}{c}{\multirow{3}{*}{$\rho$}} & \multicolumn{1}{c}{\multirow{3}{*}{Model}}  & \multicolumn{1}{c}{\multirow{3}{*}{$T$}}  & \multicolumn{3}{c}{LIVE} & \multicolumn{3}{c}{CSIQ}& \multicolumn{3}{c}{TID2013}\\
\cmidrule(r){4-6}\cmidrule(r){7-9} \cmidrule(r){10-12}
& & & RGO & SRCC & PLCC & RGO & SRCC & PLCC & RGO & SRCC & PLCC 
\\
\midrule
\multirow{4}{*}{$1/255$} 
& DBCNN & 10K
& 0.11 & 0.88 \gray{(0.96)} & 0.91\gray{(0.97)} 
& 0.18 & 0.74 \gray{(0.89)} & 0.68 \gray{(0.92)} 
& 0.14 & 0.62 \gray{(0.86)} & 0.75 \gray{(0.92)} 
\\
& UNIQUE & 10K
& 0.12 & 0.92 \gray{(0.97)} & 0.93 \gray{(0.97)}
& 0.19 & 0.46 \gray{(0.77)} & 0.49 \gray{(0.87)}
& 0.14 & 0.58 \gray{(0.80)} & 0.58 \gray{(0.81)}
\\
& TReS & 10K
& 0.13 & 0.85 \gray{(0.97)} & 0.89 \gray{(0.97)}
& 0.19 & 0.60 \gray{(0.91)} & 0.52 \gray{(0.91)} 
& 0.14 & 0.68 \gray{(0.93)} & 0.76 \gray{(0.93)}  
\\
& LIQE & 10K
& 0.07 & 0.93 \gray{(0.97)} & 0.93 \gray{(0.96)}
& 0.10 & 0.69 \gray{(0.87)} & 0.73 \gray{(0.87)} 
& 0.07 & 0.77 \gray{(0.80)} & 0.82 \gray{(0.80)} 
\\
\cmidrule(r){1-12}
\multirow{4}{*}{$2/255$} 
& DBCNN & 10K 
& 0.22 & 0.64 \gray{(0.96)} & 0.67\gray{(0.97)} 
& 0.32 & 0.29 \gray{(0.89)} & 0.16 \gray{(0.92)} 
& 0.25 & 0.37 \gray{(0.86)} & 0.41 \gray{(0.92)} 
\\ 
& UNIQUE & 10K 
& 0.24 & 0.47 \gray{(0.97)} & 0.47 \gray{(0.97)}
& 0.34 & -0.01 \gray{(0.77)} & -0.00 \gray{(0.87)}
& 0.25 & -0.20 \gray{(0.80)} & -0.11 \gray{(0.81)} 
\\
& TReS &  10K 
& 0.25 & 0.54 \gray{(0.97)} & 0.60 \gray{(0.97)}
& 0.36 & 0.08 \gray{(0.91)} & -0.08 \gray{(0.91)} 
& 0.23 & 0.25 \gray{(0.93)} & 0.37 \gray{(0.93)}
\\
& LIQE & 10K 
& 0.20 & 0.71 \gray{(0.97)} & 0.73 \gray{(0.96)}
& 0.22 & 0.53 \gray{(0.87)} & 0.56 \gray{(0.87)} 
& 0.12 & 0.63 \gray{(0.80)} & 0.71 \gray{(0.80)} 
\\ 
\cmidrule(r){1-12}
\multirow{4}{*}{$3/255$} 
& DBCNN & 10K 
& 0.30 & 0.32 \gray{(0.96)}& 0.36\gray{(0.97)} 
& 0.44 & -0.07 \gray{(0.89)} & -0.27 \gray{(0.92)} 
& 0.32 & 0.33 \gray{(0.86)} & 0.22 \gray{(0.92)}  \\ 
& UNIQUE & 10K 
& 0.33 & 0.01 \gray{(0.97)} & -0.07 \gray{(0.97)}
& 0.45 & -0.31 \gray{(0.77)} & -0.37 \gray{(0.87)}
& 0.33 & -0.41 \gray{(0.80)} & -0.32 \gray{(0.81)}  \\ 
& TReS &  10K 
& 0.34 & 0.14 \gray{(0.97)} & 0.20 \gray{(0.97)}
& 0.49 & -0.28 \gray{(0.91)} & -0.45 \gray{(0.91)} 
& 0.31 & -0.23 \gray{(0.93)} & -0.21 \gray{(0.93)} \\ 
& LIQE & 10K 
& 0.32 & 0.39 \gray{(0.97)} & 0.32 \gray{(0.96)}
& 0.32 & 0.26 \gray{(0.87)} & 0.23 \gray{(0.87)} 
& 0.18 & 0.21 \gray{(0.80)} & 0.24 \gray{(0.80)} 
\\ 
\cmidrule(r){1-12}
\multirow{4}{*}{$4/255$} 
& DBCNN & 10K 
& 0.37 & -0.04 \gray{(0.96)}& 0.03 \gray{(0.97)} 
& 0.53 & -0.25 \gray{(0.89)} & -0.40 \gray{(0.92)} 
& 0.35 & 0.36 \gray{(0.86)} & 0.18 \gray{(0.92)} 
\\ 
& UNIQUE & 10K 
& 0.40 & -0.26 \gray{(0.97)} & -0.32 \gray{(0.97)}
& 0.54 & -0.41 \gray{(0.77)} & -0.51 \gray{(0.87)}
& 0.37 & -0.41 \gray{(0.80)} & -0.40 \gray{(0.81)} 
\\
& TReS &  10K 
& 0.42 & -0.17 \gray{(0.97)} & -0.13 \gray{(0.97)}
& 0.59 & -0.42 \gray{(0.91)} & -0.64 \gray{(0.91)} 
& 0.37 & -0.45 \gray{(0.93)} & -0.42 \gray{(0.93)}
\\
& LIQE & 10K 
& 0.36 & 0.26 \gray{(0.97)} & 0.12 \gray{(0.96)}
& 0.36 & 0.21 \gray{(0.87)} & 0.15 \gray{(0.87)} 
& 0.23 & -0.07 \gray{(0.80)} & 0.07 \gray{(0.80)} 
\\ 
\cmidrule(r){1-12}
\multirow{4}{*}{$5/255$} 
& DBCNN & 10K 
& 0.41 & -0.21 \gray{(0.96)} & -0.16 \gray{(0.97)} 
& 0.60 & -0.29 \gray{(0.89)} & -0.55 \gray{(0.92)} 
& 0.38 & 0.36 \gray{(0.86)} & 0.12 \gray{(0.92)} 
\\ 
& UNIQUE & 10K 
& 0.46 & -0.39 \gray{(0.97)} & -0.48 \gray{(0.97)}
& 0.60 & -0.44 \gray{(0.77)} & -0.62 \gray{(0.87)}
& 0.40 & -0.43 \gray{(0.80)} & -0.44 \gray{(0.81)} 
\\
& TReS &  10K 
& 0.48 & -0.35 \gray{(0.97)} & -0.33 \gray{(0.97)}
& 0.65 & -0.49 \gray{(0.91)} & -0.73 \gray{(0.91)} 
& 0.42 & -0.51 \gray{(0.80)} & -0.50 \gray{(0.81)} 
\\
& LIQE & 10K 
& 0.38 & 0.16 \gray{(0.97)} & -0.10 \gray{(0.96)}
& 0.43 & 0.02 \gray{(0.87)} & -0.10 \gray{(0.87)} 
& 0.26 & -0.19 \gray{(0.80)} & -0.10 \gray{(0.80)} 
\\ 
\bottomrule
\vspace{0.2cm}
\end{tabular}	
\label{tab: eps}
\end{minipage}
\hfill
\renewcommand{\thetable}{S2}
\begin{minipage}[c]{1\textwidth}
\captionof{table}{\small The average performance results of our attack method among $10$ runs using different random seeds. The performance results of victim models before attacking are marked in gray.}
\vspace{-0.25cm}
\setlength\tabcolsep{2pt}
\renewcommand\arraystretch{0.75}
\centering
\scriptsize  
\begin{tabular}{c c ccc|ccc|ccc}
\toprule
 \multicolumn{1}{c}{\multirow{3}{*}{Model}} &  \multicolumn{1}{c}{\multirow{3}{*}{$T$}}  & \multicolumn{3}{c}{LIVE} & \multicolumn{3}{c}{CSIQ}& \multicolumn{3}{c}{TID2013}\\
\cmidrule(r){3-5}\cmidrule(r){6-8} \cmidrule(r){9-11}
&  & RGO & SRCC & PLCC &  RGO & SRCC & PLCC & RGO & SRCC & PLCC 
\\
\midrule
DBCNN & 10K 
& 0.30 ± 0.00 & 0.32 ± 0.01 \gray{(0.96)}& 0.36 ± 0.01 \gray{(0.97)} 
& 0.44 ± 0.00 & -0.08 ± 0.01 \gray{(0.89)} & -0.28 ± 0.01 \gray{(0.92)} 
& 0.32 ± 0.00 & 0.34 ± 0.01 \gray{(0.86)} & 0.22 ± 0.01 \gray{(0.92)}  \\ \cmidrule(r){1-11}
UNIQUE & 10K 
& 0.33 ± 0.00 & 0.01 ± 0.01 \gray{(0.97)} & -0.07 ± 0.02 \gray{(0.97)}
& 0.45 ± 0.00 & -0.31 ± 0.02 \gray{(0.77)} & -0.36 ± 0.03 \gray{(0.87)}
& 0.31 ± 0.00 & -0.41 ± 0.01 \gray{(0.80)} & -0.31 ± 0.02 \gray{(0.81)}  \\ \cmidrule(r){1-11}
TReS &  10K 
& 0.34 ± 0.00 & 0.14 ± 0.01 \gray{(0.97)} & 0.19 ± 0.01 \gray{(0.97)}
& 0.48 ± 0.00 & -0.27 ± 0.01 \gray{(0.91)} & -0.44 ± 0.00 \gray{(0.91)} 
& 0.31 ± 0.00 & -0.23 ± 0.01 \gray{(0.93)} & -0.19 ± 0.03 \gray{(0.93)} \\ \cmidrule(r){1-11}
LIQE & 10K 
& 0.31 ± 0.00 & 0.39 ± 0.01 \gray{(0.97)} & 0.32 ± 0.01 \gray{(0.96)}
& 0.32 ± 0.00 & 0.26 ± 0.01 \gray{(0.87)} & 0.23 ± 0.01 \gray{(0.87)} 
& 0.19 ± 0.00 & 0.19 ± 0.02 \gray{(0.80)} & 0.22 ± 0.02 \gray{(0.80)} \\
\bottomrule
\\ 
\end{tabular}	
\label{tab: seed}
\end{minipage}
\vspace{-0.1cm}
\end{table}
We conduct additional experiments to study the attack performance of our method under different distortion thresholds and verify its stability and reliability by using different random seeds.

\section{Attack Performances with Different Distortion Thresholds}
\label{sec: threshold}
In paper, we set the distortion threshold $\rho$ to be $3/255$ in order to preserve the visual semantics and quality of adversarial examples, which is already verified by existing studies. Here, we evaluate the attack performance of our method using other different $\rho$ to confirm this choice. 
Particularly, the $\rho$ is set to be $1/255$, $2/255$, $4/255$, and $5/255$ for experiments, respectively. This is realized by setting the sampling values $\{-3/255, 3/255\}$ used in the initialization and perturbation stages to be specific $\{-\rho, \rho\}$. The experimental results are presented in Table \ref{tab: eps}, along with the performance of $\rho=3/255$.
From the results, we can observe that smaller
\par \ \par \vspace{12.5cm}
\noindent $\rho$ (i.e., $<3/255$) can better meet the quality-preserving requirement, but its performance is not competitive.
Larger $\rho$ (i.e., $>3/255$) leads to better attack performance, but it would degrade the visual quality of adversarial examples. Thus, $\rho=3/255$ is the best choice in terms of preserving the visual quality while achieving good attack performance.    

\section{Stability of the Proposed Attack Method}
We investigate the stability of our attack method by running experiments using different random seeds.
Specifically, our method is independently executed $10$ times with different random seeds against the four NR-IQA models on the three datasets. The average performance results with standard deviations are reported in Table \ref{tab: seed}. 
We can see that the average performances of our attack on three datasets highly match the results reported in paper (i.e., Table $1$), and the associated standard deviations are very small. This demonstrates the favorable stability and reliability of the proposed attack.